\pdfoutput=1
\documentclass[conference]{IEEEtran}

\usepackage{graphicx}
\usepackage{siunitx}
\usepackage{natbib}
\usepackage{booktabs}
\usepackage{makecell}
\usepackage{longtable}
\usepackage{lscape}
\usepackage{listings}
\usepackage[table]{xcolor}  % Load xcolor package
\usepackage{hyperref}
\definecolor{lightgray}{gray}{0.9}
\title{Mapping News Narratives Using LLMs and Narrative-Structured Text Embeddings}

\author{Jan Elfes \\
  University College Dublin\\
  \texttt{jan.elfes@ucd.ie}}

\begin{document}
\maketitle

\begin{abstract}
  Given the profound impact of narratives across various societal levels, from personal identities to international politics, it is crucial to understand their distribution and development over time. This is particularly important in online spaces. On the Web, narratives can spread rapidly and intensify societal divides and conflicts. While many qualitative approaches exist, quantifying narratives remains a significant challenge. Computational narrative analysis lacks frameworks that are both comprehensive and generalizable. To address this gap, we introduce a numerical narrative representation grounded in structuralist linguistic theory. Chiefly, Greimas’ Actantial Model represents a narrative through a constellation of six functional character roles. These so-called actants are genre-agnostic, making the model highly generalizable. We extract the actants using an open-source LLM and integrate them into a Narrative-Structured Text Embedding that captures both the semantics and narrative structure of a text. We demonstrate the analytical insights of the method on the example of 5000 full-text news articles from Al~Jazeera and The Washington Post on the Israel-Palestine conflict. Our method successfully distinguishes articles that cover the same topics but differ in narrative structure.
\end{abstract}

\section{Introduction}
Over the past decades, \textit{narrative} has made its way into many different fields from psychology \citep{sarbin_narrative_1986}, to sociology \citep{maines_narratives_1993}, cognitive sciences \citep{herman_narrative_2001}, and economics \citep{shiller_narrative_2019}. Narratives serve as influential forces shaping individual opinions and identities, but also group-level developments and entire economies. Narratives wield significant influence on large social developments such as climate change \citep{bushell_strategic_2017}, democratic elections \citep{polletta_deep_2017}, and the coronavirus pandemic \citep{de_saint_laurent_internet_2021}.

Narratives are also highly relevant in conflict, supporting opposing group identities \citep{bar-tal_sociopsychological_2014}. With international news coverage and social media, such narratives reach far beyond the conflict zone, influencing international attention and support for the involved factions. For instance, labeling the invasion of Ukraine as such versus referring to it as a Russian military operation, can significantly alter perceptions of the events. Similarly, in the case of Israel and Palestine, perspectives shift easily depending on who is seen as having started or provoked the conflict, perpetuating a cycle of retaliation. 

Recent research has started to analyze narratives on a larger scale and the field of Computational Narrative Understanding has emerged. Related work focuses on defining narrative for computational applications and translating this definition into practice. \citet{piper_narrative_2021} formalize narrative along the categories of agents, events, temporality, setting, and perspective and review individual contributions within these categories. Further, they point out avenues of future research to combine them into a comprehensive analysis of narrative. Other related surveys focus on fictional narratives and character network representations \citep{labatut_extraction_2020}, fragmented narratives across social media posts  \citep{ranade_computational_2022}, and combining computational approaches with linguistic theory \citep{santana_survey_2023}.

Shifting towards application, two main approaches exist: deductive and inductive. \citet{coan_computer-assisted_2021} developed a BERT-based classifier to identify contrarian narratives about climate change in blogs and articles. Their approach is deductive in that it relies on a predefined taxonomy of narratives and labeled training data, making it challenging to generalize to other topics.

Most other contributions focus on the extraction of actors and their relations \citep{chambers_unsupervised_2008, mohr_graphing_2013, chaturvedi_unsupervised_2017, tangherlini_automated_2020, bandeli_framework_2020}. For example, \citet{ash_relatio_2024} use Semantic Role Labelling \citep{gildea_automatic_2002} to extract agents and patients connected by an action. This results in narrative statements such as ``people lose job'' \citep[p. 122]{ash_relatio_2024}. Although these inductive approaches are not reliant on a predefined label set, they do not offer a comprehensive structure that enables the contrast and comparison of narratives across diverse sources and topics.

We aim to advance computational narrative analysis by providing a comprehensive and generalizable narrative representation suitable to analyze and compare narrative trends across a diverse range of topics. We ground our model in structuralist linguistic theory. In particular, we use Greimas' Actantial Model, which classifies a narrative via six functional character roles: Subject, Object, Sender, Receiver, Helper, and Opponent. These so-called actants are further organized along the three fundamental relationships of desire, communication, and power \citep{greimas_structural_1984}. Extending previous work on folktales by \citet{propp_morphology_1968}, Greimas aimed to develop a general framework that can represent a narrative irrespective of the genre, which aligns with our objective.

We extract Greimas' actants from text using zero-shot prompting, an approach that has shown some success when applied to Propp's characters and elementary functions in folktales \citep{stammbach_heroes_2022, gervas_tagging_2024}. Once the Actantial Model is extracted, we convert the actants into numerical vectors using text embeddings. These vectors are then combined to create a narrative representation that captures both semantic content and narrative structure. By analyzing this representation space, we can identify narrative trends as clusters within it. We interpret these clusters as \textit{cultural narratives}, which are overarching patterns that link various narrative discourses within a community, offering members a framework for understanding and interpreting events \citep{baier_narratives_2023}.

We demonstrate the approach on a corpus of 5342 news articles on the Israel-Palestine conflict, sourced from Al Jazeera and The Washington Post. The data contains over one year of news coverage including the Hamas attack on October 7, 2023, and several months of Israel's military response. Clustering the narrative representations of all articles we identify 18 distinct narrative trends that shape the coverage of the conflict. Our analysis reveals a clear distinction in the editorial narratives between the two sources, as well as some overarching themes that persist across both outlets.

The rest of the paper is structured as follows. Section~\ref{sec:related_work} provides an overview of related research. Section~\ref{sec:narrative_theory} introduces the narrative theory in more detail. Section~\ref{sec:methodology} explains the methodology. Section~\ref{sec:data} describes the dataset. Section~\ref{sec:results} provides the analysis of the case study and section~\ref{sec:conclusion} concludes.
\section{Related work}\label{sec:related_work}
Before delving into the complexities of narratives, topic modeling has been a common task in Natural Language Processing. Arguably, the most widely recognized approach is Latent Dirichlet Allocation \citep{blei_latent_2003}, which identifies topics as collections of frequent terms. More recently, BERTopic has advanced topic modeling by leveraging BERT-based text embeddings to identify the most central topics \citep{grootendorst_bertopic_2022}. However, narrative analysis goes a step further by examining how a particular topic is portrayed.

Distinguishing narratives from frames is a long-fledged debate in the social sciences that is still ongoing \citep{aukes_narrative_2020}. In his influential work, \citet{entman_framing_1993} describes frames via their characteristic to increase or decrease the salience of certain issues. We distinguish between frames and narratives according to their complexity, i.e., a narrative can be compressed into a frame but not vice versa. Most work on frames focuses on building a classifier for a predefined label set (see \citealp{ali_survey_2022} for an overview). That said, we acknowledge that with our focus on more abstract narratives, we enter a domain where the distinction between frames and narratives becomes increasingly blurred. Some researchers have already called for a new category \textit{narrative-frames} to describe a similar viewpoint \citep{reiter-haas_computational_2024}.

% We use \textit{Llama-3-8B-Instruct} to power the extraction of actants.
% While GPT-3 \citep{brown_language_2020} and its newer siblings have revolutionized the usage and application of generative LLMs in the wider public, concerns about transparency have hampered their use in scientific applications. As a result, open-source alternatives like Llama \citep{touvron_llama_2023} and Mistral \citep{jiang_mistral_2023} have found traction. Especially smaller versions with less than 8B parameters strike a beneficial balance between performance and resource efficiency that makes them accessible to many researchers. 

\section{Narrative theory}\label{sec:narrative_theory}
In this section, we describe the underlying narrative theory of our approach. First, we introduce the overarching concept of cultural narrative. Second, we describe how we formalize narrative on the article level for our computational framework.

\subsection{Cultural Narrative}
Many definitions of narrative exist, each foregrounding different aspects of the concept. In this work, we refer to the notion of \textit{cultural narrative}. The term has been recently introduced by \citet{baier_narratives_2023}, but previous work has described similar phenomena as collective narratives \citep{bliuc_cooperation_2022} or masterplots \citep{abbott_rhetoric_2008}. In essence, a cultural narrative abstracts from individual stories to focus on shared meanings and overarching trends that link various narrative discourses within a community. It serves as an interpretative framework, helping us understand new information by identifying common themes that represent collective beliefs. We apply this concept to interpret clusters of articles, where each article contributes a unique narrative, but our interest lies in uncovering the overarching themes.

\subsection{The Actantial Model}
To structure the narrative representation of individual articles, we rely on Greimas' Actantial Model \citep{greimas_structural_1984}. While cultural narratives emerge from a collection of articles, the Actantial Model provides the necessary framework to analyze the narrative elements within each article. The Actantial Model consists of six actants, i.e., Subject, Object, Sender, Receiver, Helper, and Opponent (see Figure~\ref{prompt:actantial_model}). Actants are syntactic categories defined by their function within the narrative and are structured around three fundamental relationships. The Subject \textit{desires} the Object, which is in turn \textit{communicated} from the Sender to the Receiver. The Helper and Opponent exert \textit{power} on the Subject. A specific narrative is represented by its semantic investment, that is, the manifestation of each actant as a concrete actor (or object) in the narrative. For example, a common narrative we encounter in our data is Israel (Subject and Sender) desiring the Gaza Strip and directing its actions toward it (Object and Receiver). The US supports Israel in its desire, while Hamas opposes it.

This framework abstracts from previous work on folktales \citep{propp_morphology_1968} to a more general model. Together with the functional definition of the actants, this allows us to compare narratives across genres, topics, and media. Therefore, in theory, our framework can track narratives as they move across news articles, readers' comments, social media posts, and political speeches. Moreover, identifying the six actants aligns with established machine learning paradigms, such as Named Entity Recognition, building on the extensive prior work in this field.

Beyond its simple appearance, the Actantial Model can capture complex information in its structure. An example is the \textit{actant syncretism}, which describes the amalgamation of two actants in one actor. An example from a folktale is the Subject-Receiver syncretism, where the hero both desires and ultimately receives the Object at the end of their journey. Therefore, a syncretism structurally captures nuances of the narrative, in this case, the hero's success. 

\section{Methodology}\label{sec:methodology}
\begin{figure}[t]
    \centering
    \includegraphics[width=0.475\textwidth]{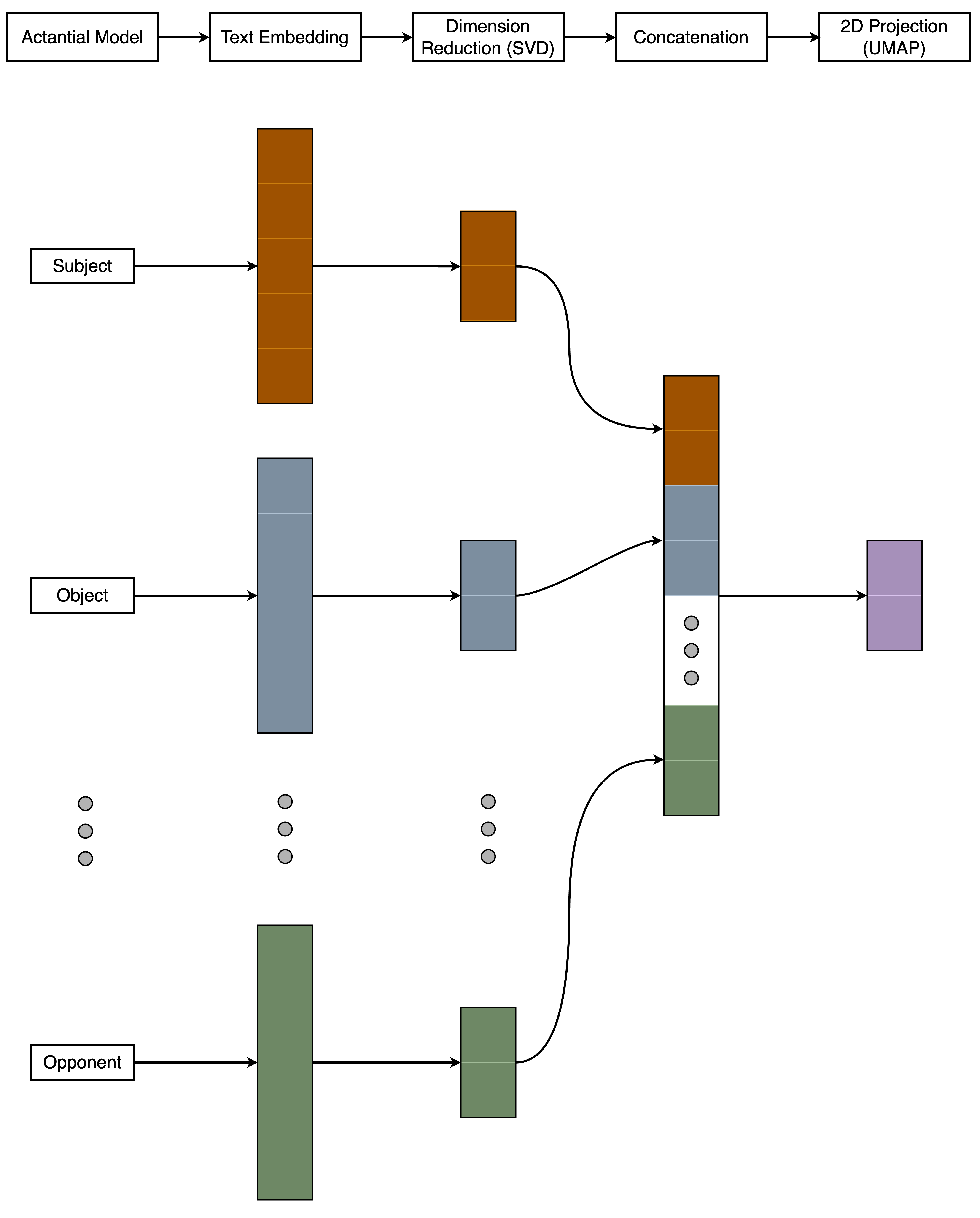}
    \caption{Schema of the embedding process. Each actant is embedded using text embeddings that are then reduced in dimension using SVD. Next, we concatenate the reduced embeddings into the narrative-structured embedding. Lastly, we project the embedding to a plane for visualization and clustering.}
    \label{fig:narrative_embedding}
\end{figure}

Figure \ref{fig:narrative_embedding} sketches our extraction pipeline from individual news articles to clusters of cultural narratives. We start by prompting a generative LLM with the extractions to provide the Actantial Model for a given article. The output of the model is a list of six actors that realize the positions of the Actantial Model within the article. We then translate this natural language representation into a set of six numerical vectors using text embeddings. After reducing the dimension of each embedding, to reduce variance, we concatenate them into one vector. This narrative-structured text embedding provides a representation of the article that is sensitive to both actant semantics and narrative structure. Lastly, we project the embeddings of our corpus onto a plane using UMAP. This process emphasizes the cluster structure in our data, allowing us to identify groups of articles sharing a cultural narrative. The individual steps are described in detail in the following sections.

\subsection{Prompting}
The initial step is extracting the Actantial Model, i.e., Subject, Object, Sender, Receiver, Helper, and Opponent from each news article. We prompt a decoder-only LLM with specific instructions to provide a suitable output. We take inspiration from \citet{wang_gpt-ner_2023} who adapted GPT for Named Entity Recognition. The prompt includes the actant label set, a brief definition for each actant, the full-text news article, and an instruction to extract the actants in JSON format (see Figure \ref{prompt:actantial_model}). 

\begin{figure*}
\begin{lstlisting}
According to the Actantial Model by Greimas with the actant label set ["Sender", "Receiver", "Subject", "Object", "Helper", "Opponent"], the actants are defined as follows:

* Subject: The character who carries out the action and desires the Object.
* Object: The character or thing that is desired.
* Sender: The character who initiates the action and communicates the Object.
* Receiver: The character who receives the action or the Object.
* Helper: The character who assists the Subject in achieving its goal.
* Opponent: The character who opposes the Subject in achieving its goal.

Based on this Actantial Model and the actant label set, please recognize the actants in the given article.

Article: {{ article }}

Question: What are the main actants in the text? Provide the answer in the following JSON format: {"Actant Label": ["Actant Name"]}. If there is no corresponding actant, return the following empty list: {"Actant Label": []}.

Answer:
\end{lstlisting}
% \captionsetup{labelformat=prompt, labelsep=colon}
\caption{Prompt used to extract the Actantial Model from a news article. During inference \texttt{\{\{ article \}\}} is replaced with a full-text news article.}
\label{prompt:actantial_model}
\end{figure*}

\subsubsection{Model specifications}\label{sec:model_specification}
All prompting experiments use \textit{Llama-3-8B-Instruct}, the smaller of the two currently publicly available models from Meta's Llama 3 family \citep{dubey_llama_2024_ed}. Unlike the base version, the \textit{Instruct} model is fine-tuned to follow natural language prompts. The model is state-of-the-art with a competitive performance compared to popular closed-source models like GPT3.5. Furthermore, with 8B parameters, the model strikes a good balance between performance and resource efficiency.

We run model inference without sampling, meaning the model selects the highest probability token at each step. As a result, all experiments are deterministic and reproducible. Moreover, the model is open-source, ensuring full access to the model parameters. Additional model specifications and the code for our extraction pipeline are available on GitHub\footnote{\url{https://github.com/jelfes/llam}}. We run our experiments using Nvidia L40S (48GB) and Nvidia Tesla A100 (40GB) GPUs.

\subsection{Text embeddings}
After extracting the Actantial Model, we translate its natural language representation into a numerical form. Text embeddings transform the semantic content of a phrase into a position in high-dimensional space. Therefore, semantically similar words and phrases are placed close in this space, with distance defined by the cosine similarity. The most common embedding models are adaptations of BERT \citep{devlin_bert_2019}. Most commonly, variants of Sentence-BERT \citep{reimers_sentence-bert_2019}. Recently, also decoder-only LLMs have been adapted for text-embedding tasks \citep{wang_improving_2024, behnamghader_llm2vec_2024}. Though these models achieve state-of-the-art results, they are much larger and therefore more expensive than BERT-based models with only a small increase in performance on the MTEB benchmark \citep{muennighoff_mteb_2023}.

We chose the more lightweight BERT-based \textit{E5-large} with an embedding dimension of 1024 \citep{wang_improving_2024}. With an average score of $61.42$, \textit{E5-Large} currently ranks as one of the most lightweight options in the top 70 of the MTEB leaderboard \citep{mteb_mteb_2024}. With only 335M parameters, and being easily accessible via Huggingface, the model combines performance, computational efficiency, and convenience for our use case. Additionally, in our initial experiments, its performance surpassed that of the popular \textit{MiniLM} \citep{wang_minilm_2020} which serves as default text-embedding on Huggingface.

\subsection{Narrative-structured text embeddings}
To create a narrative-structured text embedding, we combine the individual text embeddings of the six actants into one vector, thus representing the full text through the semantics and structure of its Actantial Model (Figure \ref{fig:narrative_embedding}). We concatenate the text embeddings into a 6x1024-dimensional vector. Compared to other methods like mean pooling, concatenation creates an embedding sensitive to both structural and semantic changes. For example, exchanging the Subject with a new actor will change the embedding according to their semantic difference. Further, switching the Subject and the Object will change the embedding due to structural differences, albeit maintaining the overall semantics.

\subsection{Dimension reduction}\label{sec:dim_reduction}
Concatenating the word embeddings has one major drawback, the dimension. Identifying clusters in a high-dimensional space is difficult as distances between points and clusters become less significant, i.e., the curse of dimensionality. The common solution is dimension reduction. 

Moreover, our method is designed to analyze unknown data and therefore has no fixed label set. This means the model provides different terms such as “Netanyahu”, or “Benjamin Netanyahu” to refer to the same actor. Text embeddings solve this issue to some extent as the embeddings for both actors will be close due to their semantic similarity. Reducing the dimension of the actor embeddings can further minimize nuanced differences in wording, thereby simplifying the clustering process.

We apply what we call micro dimension reduction. Instead of reducing the dimension of the whole concatenated embedding, we apply Singular Value Decomposition (SVD) to each of the six actant components. Previous work by \citet{zhang_evaluating_2024} has identified SVD as a robust method to reduce the dimension of text embeddings. SVD identifies the axis with the highest variance, allowing us to emphasize distinguishing features rather than minor nuances between similar actants. We reduce the dimension from 1024 to 34, which strikes a good balance between reducing variance and preserving important distinctions between actors. We provide additional details in section  \ref{sec:ab_dim_reduction}. Therefore, the final \textit{SVD-~34} embedding has the dimension $6\times34=204$.

To further enhance cluster resolution, we apply UMAP \citep{mcinnes_umap_2020} at a macro level, projecting the \textit{SVD-~34} embeddings onto a plane. Unlike SVD, UMAP is not linear. Instead, it constructs a graph by connecting points to their neighbors and iteratively adjusts the distances, bringing closer points together and pushing farther points apart. This makes it a popular preprocessing step for both clustering and visualizing high-dimensional data. 

\subsection{Clustering}
Finally, we cluster the data to identify cultural narratives that span across articles. We apply Agglomerative Clustering with ward linkage to the 2-dimensional UMAP representation of the data. The hierarchical clustering paradigm follows the intuition that narratives can appear in layers. With different levels of granularity, larger narratives split up into smaller variants. We identify the optimal clustering using the silhouette score. 

\section{Data}\label{sec:data}
\begin{figure}
    % \centering
    \includegraphics[width=0.475\textwidth]{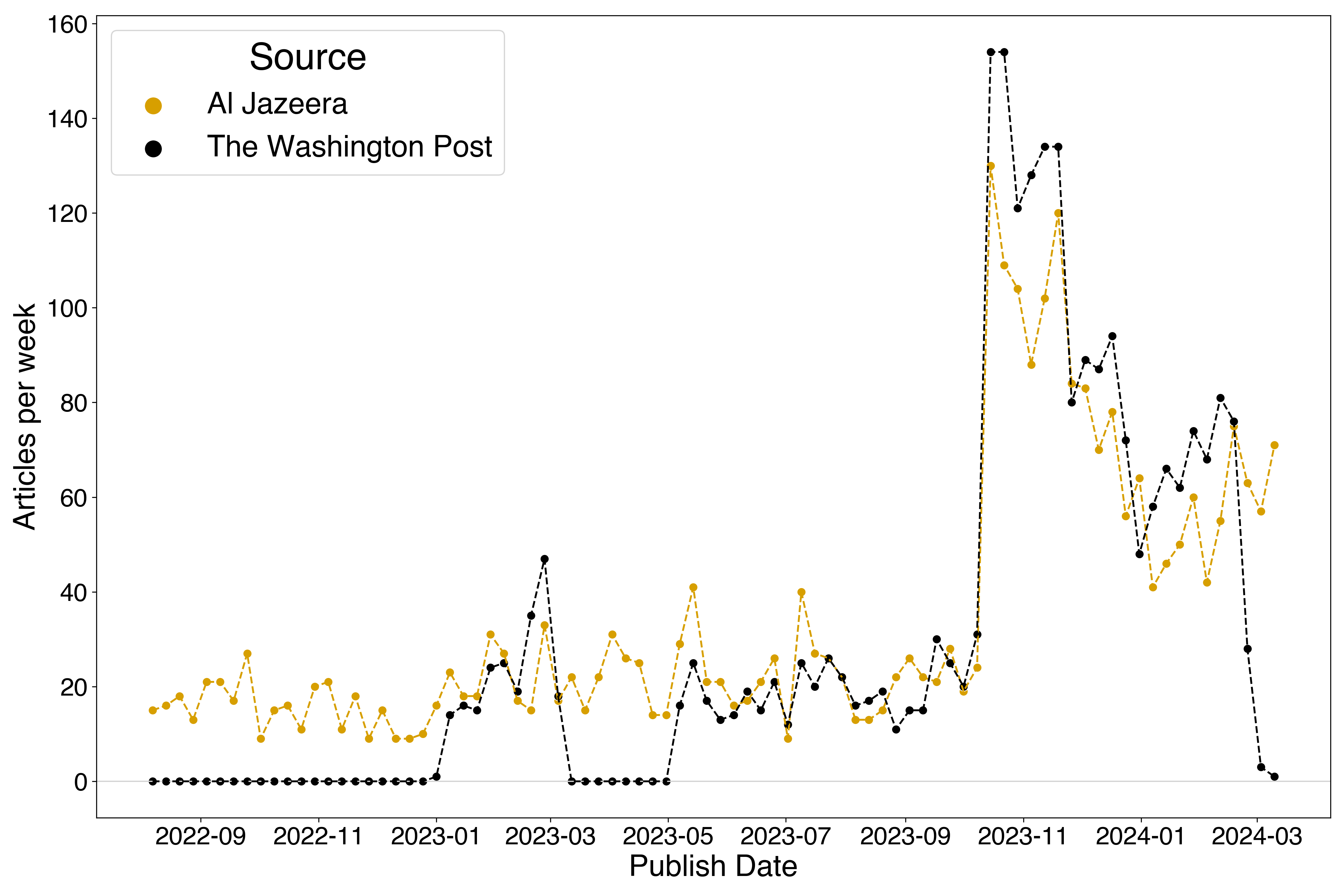}
    \caption{Number of articles per week for Al Jazeera and The Washington Post. With a total number of 5342 articles.}
    \label{fig:timeline_all_articles}
\end{figure}

The dataset comprises 5,342 full-text news articles. Al Jazeera English and The Washington Post were selected as sources, due to their accessibility, English-language content, international reach, and distinct editorial narratives. Using Media Cloud \citep{roberts_media_2021}, we compiled the URLs for every article published by either source between August 1, 2022, and March 10, 2024. Next, we built a scraper to extract the full-text articles from each URL\footnote{\url{https://github.com/jelfes/NewsScraper}}. The scraper collects all text that is enclosed in HTML paragraph tags. This proved effective and robust without the need for any post-processing of the data. We selected articles that contain at least one of ``Israel'', ``Palestine'', ``Gaza'', or``Hamas''. Further, we only selected Al Jazeera articles from the ``news'' section. Articles from Al Jazeera account for \SI{54}{\percent} of the final dataset. The distribution of articles over time is shown in Figure \ref{fig:timeline_all_articles}. The distribution of article length is shown in Figure \ref{fig:word_count}. The Washington Post articles are longer on average with 1109 words compared to Al Jazeera with 699 words.

\section{Results}\label{sec:results}
We applied the narrative extraction pipeline outlined in section~\ref{sec:methodology} to the Israel-Palestine data (section~\ref{sec:data}). Due to the length of the articles, it is not unlikely to identify more than one actor for each actant. However, for simplicity, we focused on the first actor extracted. Table \ref{tab:top_3_actors} summarises the most common actors divided by news source. Israel is the main Subject and Gaza is the main Object for both Al~Jazeera and The Washington Post. The remaining actors indicate the different editorial narratives of each source. While The Washington Post focuses on the US as the main Helper and Hamas as the main Opponent, Al~Jazeera places Israel as the main Opponent and Qatar and Egypt as Helpers.

\begin{figure*}
    \centering
    \makebox[\textwidth][c]{\includegraphics[width=1\textwidth]{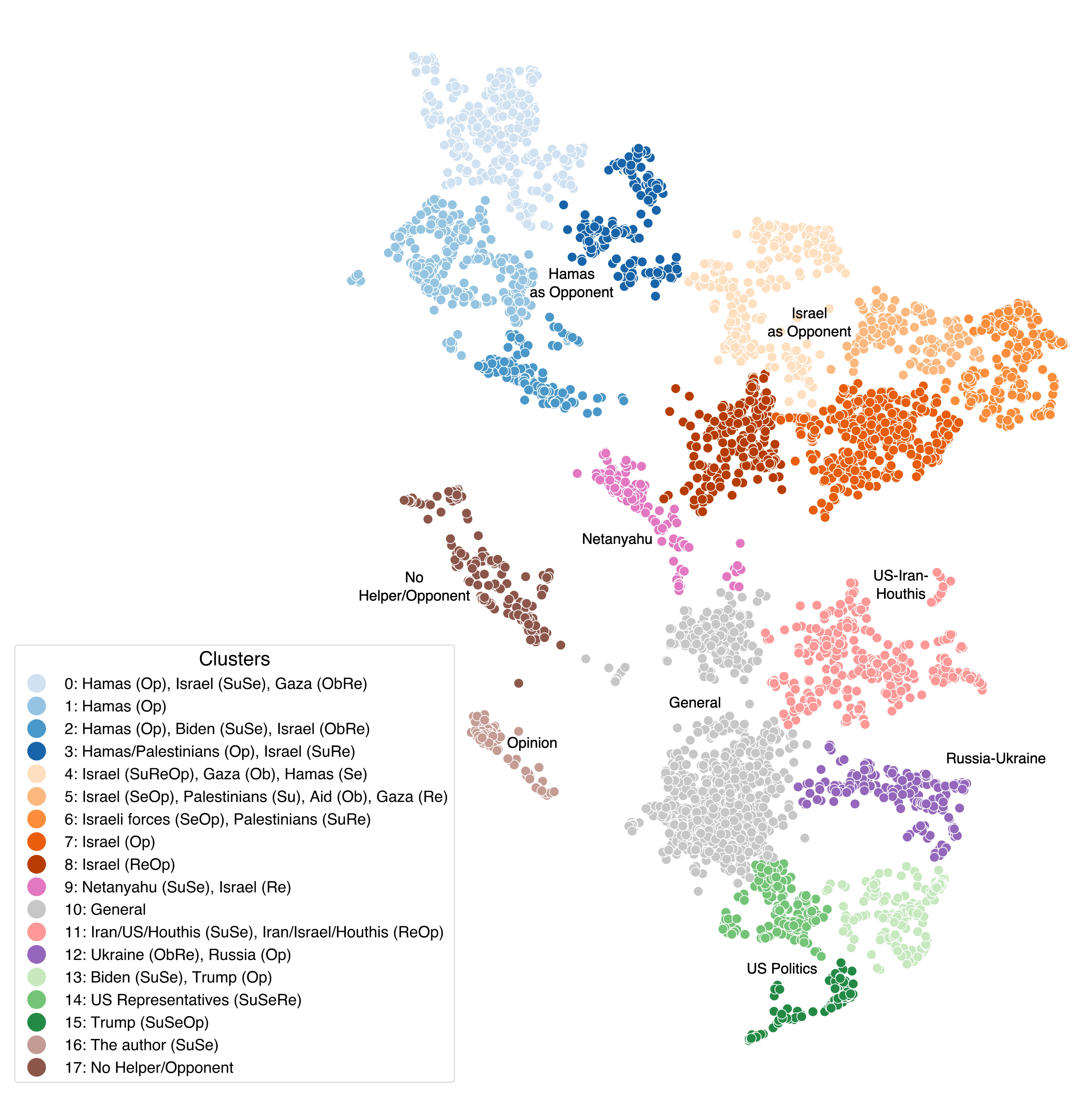}}%
    \caption{Map of 18 clusters identified in the Israel-Palestine news coverage. Each marker represents one article. The text annotations indicate the general topic of the respective area of the map.}
    \label{fig:umap_clusters}
\end{figure*}

We identified the optimal clustering using the silhouette score. Of these 20 clusters, we dropped one as an outlier and merged the two most central clusters into one as they did not provide a clear narrative trend. This left us with 18 clusters, shown in Figure \ref{fig:umap_clusters}. The clusters are labeled via their main actors. Each label contains the most common actor per actant if they occur in at least \SI{20}{\percent} of articles. We state the name of the actor followed by the initials of the actantial roles it occupies. For example “Israel (SuSe)”  indicates that Israel is both the most common \textbf{Su}bject and \textbf{Se}nder in that cluster. The order of the actants is chosen freely for emphasis. Table \ref{tab:clusters} contains additional details, listing the three most common actors per actant and cluster. To improve readability, we only include actors that occur in at least \SI{5}{\percent} of articles in the cluster.

Before analyzing the clusters, a few considerations on UMAP and the interpretation of the cluster locations. Based on text embeddings and the Actantial Model, the initial narrative representation translates the narrative structure and actant semantics into a position in a high-dimensional space. Semantically similar expressions have high cosine similarity and, thus, are close to each other. This characteristic is mostly preserved when using SVD. The UMAP algorithm amplifies the original distances by bringing closer points together and pushing farther points apart.  Therefore, when interpreting the plot, one should focus on the relative position of the points and the clusters, not the absolute. Further, although UMAP preserves both local and global structure, the interpretability of the relative position of the clusters decreases with distance.

\subsection{Cluster overview}
We identified two large components containing 9 of the 18 clusters in Figure \ref{fig:umap_clusters}. In the upper left corner, shaded in blue, we find 4 related clusters all sharing Hamas as the main Opponent. The 5 clusters in the top right, shaded in orange, share Israel as the main Opponent (see Figure \ref{fig:narrative_map_top_half} for a closer view). 

The component ``Hamas as Opponent'' splits up into 4 different clusters. Cluster 0 focuses on Israel as the Subject and Sender and Gaza as the Object. Articles in this cluster provide a detailed account of Israel's operations in the Gaza Strip with a focus on Israel as the main driver of the action. This cluster is also the most balanced with nearly equal shares of articles from both sources (Table \ref{tab:source_shares}). The next cluster, cluster 1, has the highest share of Hamas as the main Opponent with \SI{91}{\percent} (Table \ref{tab:clusters}). The rest of the actors are more diverse. Considering a sample from the cluster, we find that it revolves around various issues, sharing Hamas as the Opponent. Examples include a report of the Spanish Foreign Minister urging differentiation between Palestinian civilians and Hamas \citep{al_jazeera_we_2023_ed}, and an account discussing antisemitic double standards towards Israel \citep{rubin_opinion_2023}. In cluster 2 US President Joe Biden is the main Subject and Sender, voicing his country's support for Israel, the Receiver, against Hamas, the Opponent. The last cluster in this component, cluster 4, provides a range of Palestinian actors as Opponents in addition to Hamas. The articles in this cluster include accounts of the October 7 attack, referring to Hamas fighters as ``Palestinian militants'' \citep{george_amid_2023}, as well as a summary of the Palestinian resistance Israel might face as it prepares for its military operation in the days following the attack \citep{kusovac_how_2023}.

It is important to note that ``Hamas as Opponent'' does not necessarily mean that these articles are ``pro-Israel''. The pro or contra sentiment is often transported via additional nuances such as emphasizing the deadly actions of one side or the other. What ties these narratives together is that they all focus on actions that Hamas opposes. For example the article ``Which of Gaza’s hospitals is Israel threatening?'' \citep{al_jazeera_hospitals_2023} from cluster 0 describes how Israel is targeting hospitals as potential Hamas bases in the Gaza Strip. Although Hamas is opposing Israel's actions, Israel is depicted in a negative light by listing civilian casualties in the affected hospitals. Conversely, ``A fraught battlespace awaits Israel after the pause'' \citep{ignatius_opinion_2023} discusses Israel's options after the end of a truce with Hamas and reconciles the horrors of the October 7 attack. Both articles share a similar cultural narrative: a country's fight against a terrorist group. The pro or contra sentiment is transported through the valence of civilian suffering on one side or the other.

The second component focuses on ``Israel as Opponent''. The first cluster in this component, cluster 4, addresses Israel's operation in Gaza. However, as opposed to cluster 0, here Israel is more often the Receiver, and Hamas the Sender. Many articles in this cluster address attacks against Israel while its troops operate in Gaza. Compared to cluster 0, the cultural narrative shifts towards the resistance against a military occupation. To the right, cluster 5 contains articles with a range of Palestinian actors as the Subject and Receiver. The Object focuses on aid, peace, and a humanitarian corridor. Articles in the upper right corner, cluster 6, emphasize the contrast between military and civilian actors with Israeli forces as Sender and Opponent and Palestinians as Subject and Receiver. Below the previous two clusters is a larger collection of articles, cluster 7, describing different forms of Israel's misconduct. This includes South Africa's appeal in front of the International Court of Justice to classify Israel's actions in Gaza as genocide \citep{parker_what_2024} as well as a range of articles on the deaths of journalists in the Gaza Strip \citep{al_jazeera_jazeera_2023_ed}. The last cluster in this component centers around Israel as the Receiver of messages from a range of international actors (cluster 8). 

The third bigger component in the lower right of the map, shaded in green, addresses ``US Politics'' (clusters 13-15). It splits up into three clusters with different actors as both Subject and Sender. Those are Joe Biden, Donald Trump, and members of the US House of Representatives. Different than in cluster 2, where Joe Biden also appeared as Subject-Sender, the Israel-Palestine conflict plays only a peripheral role in this component that is focused more on US domestic issues.

The rest of the clusters are more individual and not grouped into larger components. Cluster 9 focuses on the Israeli Prime Minister as the Subject and Sender. The main Object is a change to Israel's judicial system \citep{rubin_what_2023}. Cluster 11 is positioned between ``US Politics'' and ``Israel as Opponent'' and describes the involvement of Iran and the Houthi rebels in the conflict \citep{al_jazeera_houthis_2024_ed}. Cluster 12 deals with the Russian invasion of Ukraine and issues relating to both conflicts such as US funding for Ukraine and Israel \citep{goodwin_senate_2024}. US President Biden plays a role as Sender in this cluster, placing it closer to the ``US Politics'' component. The dense cluster in the center of the map contains a diverse mix of articles without a clear narrative trend.

To the left of these clusters, we find two separate islands. Each has a unique feature that separates it from the main body of articles. Most of the articles in ``No Helper/Opponent'' do not have a Helper (\SI{94}{\percent}) or an Opponent (\SI{73}{\percent}). In comparison, out of the whole data \SI{28}{\percent} of articles are missing a Helper, and only \SI{6}{\percent} are missing an Opponent. The high number of missing actants in this cluster is partially explained by a relatively low word count. The only other cluster with a similarly high absence of Opponents is cluster 16, positioned adjacent to this one. It is also the primary cluster containing actors not explicitly mentioned in the text, such as \textit{the author} and \textit{the reader}. Many of these articles are opinion pieces written in the first person. 

\subsection{Actantial Motifs}
Considering the structure of the Actantial Model, we identify several recurring motifs within the clusters. A common pattern is the Subject-Sender syncretism which we find in \SI{41}{\percent} of articles (Table~\ref{tab:syncretism}). The most common actor in this function is Israel (\SI{13}{\percent} of cases), followed by Joe Biden, Benjamin Netanyahu, Hamas, and Donald Trump. This configuration merges the two actants that guide the action and control the situation, positioning them in a commanding role within the narrative. Second is the Subject-Receiver syncretism, which also positions the actor as the central figure in the narrative but with limited agency. Israel and Palestinians appear mainly in this role. The Subject-Opponent syncretism occurs less frequently but is the most one-sided, with Israel accounting for \SI{68}{\percent} of the cases, primarily in articles published by Al~Jazeera. Although we previously observed that the Opponent is not always portrayed negatively, this combination suggests a tendency toward a villain narrative.

\subsection{Comparing news sources}
Given the different readerships of the two news outlets, we expect their average narratives to differ. As a major military partner, the US has consistently voiced its support for Israel following the October 7 attack. Therefore, we expect The Washington Post, a US-based newspaper, to feature more articles with narratives that support Israel. Al~Jazeera, based in Qatar, has a large Arabic-speaking audience. Although we analyze Al~Jazeera English, previous research has found the reporting to be similar to its Arabic counterpart \citep{fahmy_-jazeera_2011}. Due to the long history of the Israel-Palestine conflict and the tense relations between many Arab countries and the US, we expect to find more narratives supporting the Palestinian population in Gaza and critical reports on Israel.

All clusters contain articles from both news sources, albeit not always equally distributed (see Table \ref{tab:source_shares}). The clusters \numrange[range-phrase = --]{13}{15}, dealing with US Politics all contain over \SI{75}{\percent} articles from The Washington Post. On the other hand, the clusters \numrange[range-phrase = --]{4}{8}, that position Israel as Opponent, are mainly covered by Al~Jazeera. The most one-sided is cluster~6 with Al~Jazeera articles accounting for \SI{91}{\percent} of the content. The prevalent narrative directly contrasts Palestinian civilians with the Israeli military, highlighting a power imbalance. In contrast, The Washington Post focuses more on the direct opposition between Israel and Hamas.

\begin{figure}[ht]
    \centering
    \includegraphics[width=0.475\textwidth]{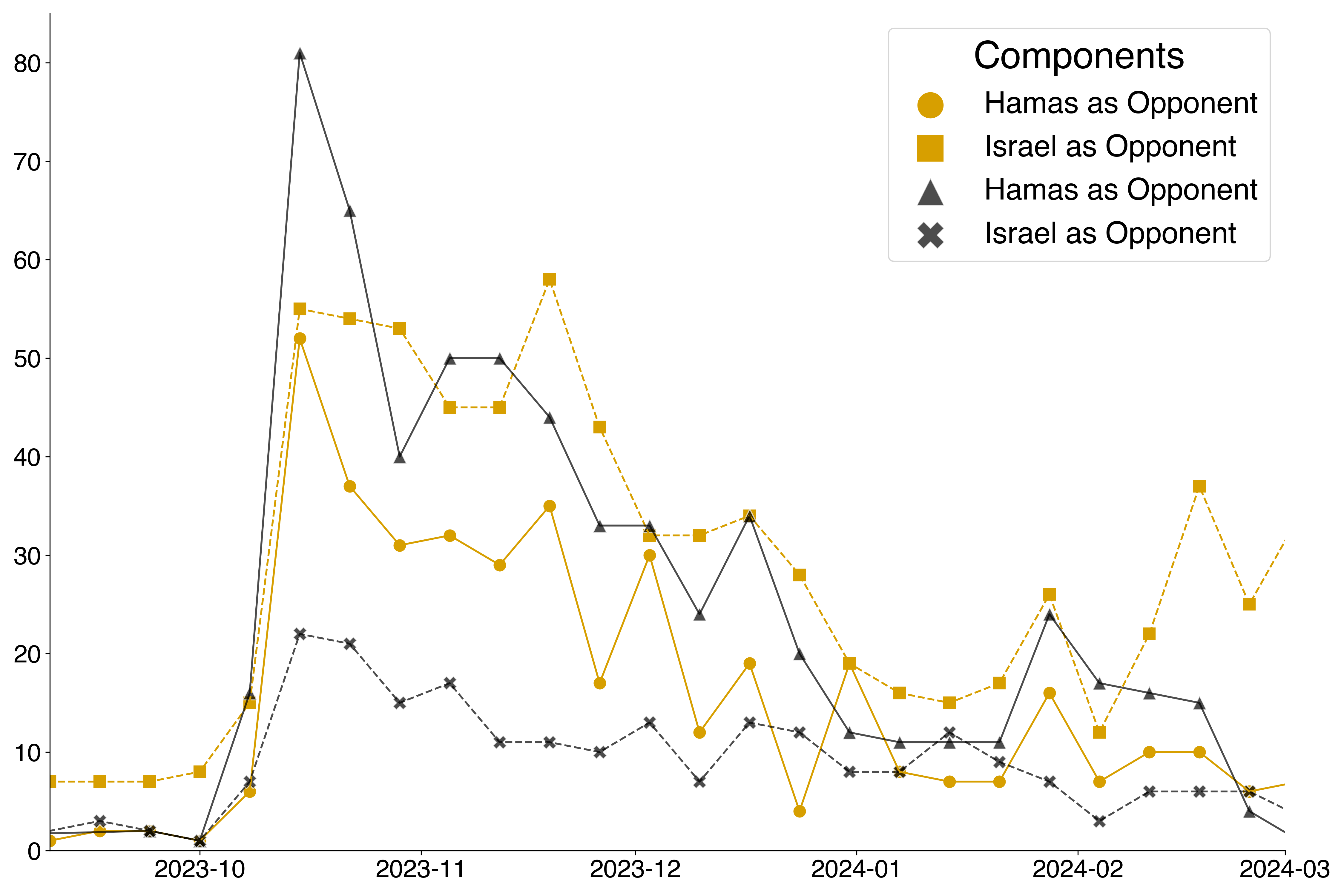}
    \caption{Number of articles per week between September 2023, and March 2024 by Al~Jazeera (yellow) and The Washington Post (black) for the components ``Hamas as Opponent'' (clusters \numrange[range-phrase = --]{0}{3}) and ``Israel as Opponent'' (clusters \numrange[range-phrase = --]{4}{8}).}
    \label{fig:timeline_components_by_media}
\end{figure}

Figure \ref{fig:timeline_components_by_media} shows the development of the two main components ``Hamas as Opponent'' (clusters \numrange[range-phrase = --]{0}{3}) and ``Israel as Opponent'' (clusters \numrange[range-phrase = --]{4}{8})) over time. The overall number of articles peaks in the week of the October 7 attack, followed by a gradual decline over the subsequent months. We separate articles from Al~Jazeera (yellow) and The Washington Post (black) to showcase their distinct trends. Most articles in The Washington Post fall under ``Hamas as Opponent,'' with overall coverage declining steadily after the Hamas attack. ``Israel as Opponent'' maintains a steady baseline, with around 10 articles per week. In Al~Jazeera, both components peak at around 50 articles in the week following October 7. In the subsequent weeks, ``Hamas as Opponent'' declines rapidly, while ``Israel as Opponent'' experiences a second spike before also declining.

The development of the individual clusters for each of the two components can be seen in Figures \ref{fig:timeline_clusters_hamas_opponent} and \ref{fig:timeline_clusters_israel_opponent}. The most prevalent narratives within ``Hamas as Opponent'' are general opposition and condemnation of Hamas, cluster 1, and coverage of Israel's military operation, cluster 0. Articles about US President Biden's support for Israel spike occasionally, cluster 2. The coverage under ``Israel as Opponent'' mainly peaks with cluster 4 after October 7. Narratives in support of Palestinians and critical of Israel increase over the following weeks and cluster 7 becomes the most prevalent narrative in this component.

\subsection{BERTopic baseline}
\begin{figure}
    \centering
    \includegraphics[width=0.475\textwidth]{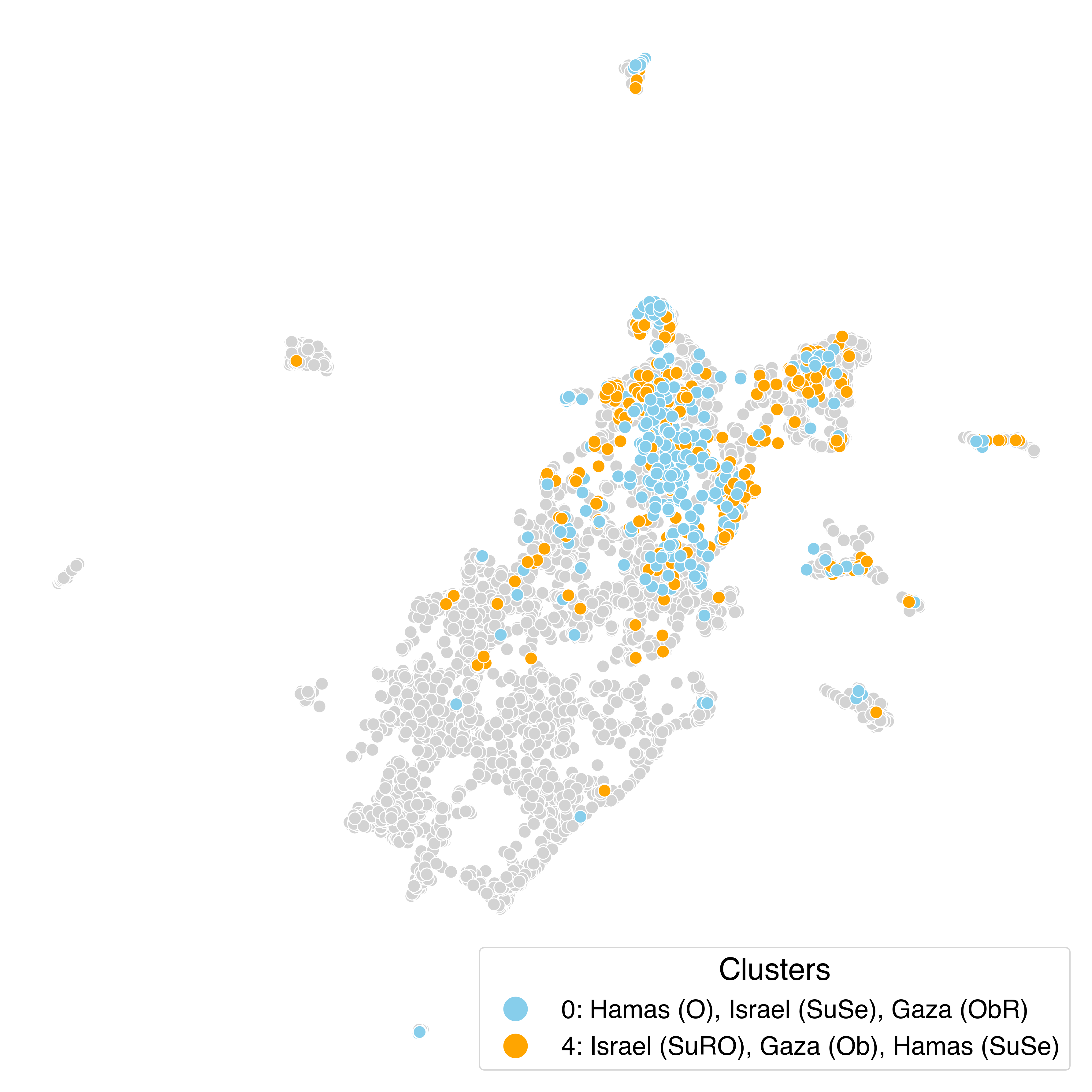}
    \caption{2D-Projection of the Israel-Palestine data using the embedding procedure of BERTopic. Colors indicate articles belonging to clusters 0 and 4, indicating a lack of separation.}
    \label{fig:bertopic}
\end{figure}

To emphasize the novel insight provided by our approach, we compare our analysis with the popular BERTopic \citep{grootendorst_bertopic_2022}. Similar to our approach, BERTopic uses text embeddings to represent articles and applies UMAP for dimensionality reduction before identifying clusters. The key difference in our method is that, instead of using a general text embedding for the entire article, we employ a narrative-structured text embedding. Figure~\ref{fig:bertopic} illustrates the distribution of articles based on BERTopic's dimension reduction procedure. The islands surrounding the central component correspond to individual topics, such as Syria, Lebanon, Netanyahu, or journalist Shireen Abu Akleh. The central cluster comprises the main set of articles that directly cover the conflict.

Clusters 0 and 4 revolve around the same set of main actors: Israel, Hamas, and Gaza. However, they differ significantly in terms of narrative structure. As shown in Figure~\ref{fig:bertopic} this difference is not captured by BERTopic. To illustrate, we compare two articles published shortly after the October 7 attack. Using BERTopic these two articles appear indistinguishable, whereas in our model, they belong to different clusters. \citet{tharoor_analysis_2023} criticizes Israel for its history of mistreating Palestinians, painting a concerning picture of the impending military response. Despite this critical stance, the article discusses the events as part of Israel's story arc as Subject and Sender. Hamas remains on the outside as the Opponent. In contrast, \citet{al_jazeera_fears_2023_ed} covers the same topic but places Hamas in a central role. By featuring interviews with Hamas officials, the article treats both actors as equally important to the story, creating a significantly different perspective.

\subsection{Limitations and future work}
We offer three considerations on the limitations of our approach. First, we confirmed the validity of the extracted actants for a random sample of articles. However, without ground-truth labels, we cannot assess the true performance of the model. Using annotated data would also enable us to compare different LLMs and identify potential biases in the models.

Second, as discussed at the beginning of this section, we only focused on the first extracted actor per actant and article. This especially limits longer news articles with a variety of actors and different narratives. 

Third, due to the open label set of the actant extraction, some actors occur in various forms, making them challenging to summarize. For example both clusters 0 and 5 feature many diverse Palestinian actors. Although this is less of an issue for the clustering algorithm, as their embeddings exhibit high cosine similarity, interpreting the final clusters remains challenging. Consequently, relying on the most common actors introduces a bias toward actors with consistent wording, e.g., names of countries and politicians. We aim to address these limitations in future work.
\section{Conclusion}\label{sec:conclusion}
We presented a narrative-structured text embedding that represents the semantics and narrative structure of a text in numerical form. The model is based on structuralist linguistic theory and can be extracted using prompting with an LLM. We used text embeddings to translate the natural language representation from the LLM into a numerical embedding. Further, we applied dimension reduction to create a more compact embedding.

We identified 18 clusters in our corpus of news articles on the Israel-Palestine conflict. Using the Actantial Model, we were able to distinguish clusters not only by their semantics but also by their narrative structure. For instance, clusters 0 and 4 revolve around the same main actors, but with different actantial roles. Moreover, we applied the concept of actant syncretism to analyze the positioning of individual actors within the narratives, identifying similarities among political figures and differences in the portrayal of Israel, Palestinians, and Hamas.

The main two cultural narratives we could identify are the fight against terrorism (clusters \numrange[range-phrase = --]{0}{3}) and the opposition against military occupation (clusters \numrange[range-phrase = --]{4}{8}). These broader trends further divide into smaller narratives, such as US support for Israel (cluster 2) and attacks against civilians (cluster 6). While all clusters contain articles from both sources, some are significantly skewed. Most notably cluster 6 contains mostly articles from Al Jazeera, showcasing its focus on civilian suffering. In contrast, cluster 0, which focuses on Israel as the central actor and descriptions of its operation in Gaza, contains an equal share from both sources. Overall, our findings align with the expected narratives in the conflict. The US-based Washington Post emphasizes US-related issues and adopts a more supportive stance toward Israel, while Al Jazeera portrays Israel more negatively, with a greater focus on Palestinian civilians.  

Our model proved effective in discerning different narratives according to their structure. Further, we provide a comprehensive and generalizable framework that is suitable for comparing narratives across different topics and media.

\section*{Acknowledgements}
The research conducted in this publication was funded by the Irish Research Council under grant number [GOIPG/2023/4198].

% \bibliography{ms.bbl}

\begin{thebibliography}{59}

%%% the Web address.

\ifx \showCODEN    \undefined \def \showCODEN     #1{\unskip}     \fi
\ifx \showDOI      \undefined \def \showDOI       #1{#1}\fi
\ifx \showISBNx    \undefined \def \showISBNx     #1{\unskip}     \fi
\ifx \showISBNxiii \undefined \def \showISBNxiii  #1{\unskip}     \fi
\ifx \showISSN     \undefined \def \showISSN      #1{\unskip}     \fi
\ifx \showLCCN     \undefined \def \showLCCN      #1{\unskip}     \fi
\ifx \shownote     \undefined \def \shownote      #1{#1}          \fi
\ifx \showarticletitle \undefined \def \showarticletitle #1{#1}   \fi
\ifx \showURL      \undefined \def \showURL       {\relax}        \fi
% invisible to TeX
\providecommand\bibfield[2]{#2}
\providecommand\bibinfo[2]{#2}
\providecommand\natexlab[1]{#1}
\providecommand\showeprint[2][]{arXiv:#2}

\bibitem[Abbott(2008)]%
        {abbott_rhetoric_2008}
\bibfield{author}{\bibinfo{person}{H.~Porter Abbott}.} \bibinfo{year}{2008}\natexlab{}.
\newblock \showarticletitle{The rhetoric of narrative}.
\newblock In \bibinfo{booktitle}{\emph{The {Cambridge} {Introduction} to {Narrative}} (\bibinfo{edition}{2} ed.)}, \bibfield{editor}{\bibinfo{person}{H.~Porter Abbott}} (Ed.). \bibinfo{publisher}{Cambridge University Press}, \bibinfo{address}{Cambridge}, \bibinfo{pages}{40--54}.
\newblock
\urldef\tempurl%
\url{https://doi.org/10.1017/CBO9780511816932.006}
\showDOI{\tempurl}


\bibitem[{Al Jazeera}(2023a)]%
        {al_jazeera_jazeera_2023_ed}
\bibfield{author}{\bibinfo{person}{{Al Jazeera}}.} \bibinfo{year}{2023}\natexlab{a}.
\newblock \showarticletitle{Al {Jazeera} journalist {Samer} {Abudaqa} laid to rest in southern {Gaza}}.
\newblock \bibinfo{journal}{\emph{Al Jazeera}} (\bibinfo{date}{Dec.} \bibinfo{year}{2023}).
\newblock
\urldef\tempurl%
\url{https://www.aljazeera.com/news/2023/12/16/al-jazeera-journalist-samer-abudaqa-laid-to-rest-in-southern-gaza}
\showURL{%
\tempurl}


\bibitem[{Al Jazeera}(2023b)]%
        {al_jazeera_fears_2023_ed}
\bibfield{author}{\bibinfo{person}{{Al Jazeera}}.} \bibinfo{year}{2023}\natexlab{b}.
\newblock \showarticletitle{Fears of a ground invasion of {Gaza} grow as {Israel} vows ‘mighty vengeance’}.
\newblock \bibinfo{journal}{\emph{Al Jazeera}} (\bibinfo{date}{Oct.} \bibinfo{year}{2023}).
\newblock
\urldef\tempurl%
\url{https://www.aljazeera.com/news/2023/10/7/world-is-watching-fears-grow-of-a-massive-gaza-invasion-by-israel}
\showURL{%
\tempurl}


\bibitem[{Al Jazeera}(2023c)]%
        {al_jazeera_hospitals_2023}
\bibfield{author}{\bibinfo{person}{{Al Jazeera}}.} \bibinfo{year}{2023}\natexlab{c}.
\newblock \showarticletitle{Which of {Gaza}’s hospitals is {Israel} threatening?}
\newblock \bibinfo{journal}{\emph{Al Jazeera}} (\bibinfo{date}{Nov.} \bibinfo{year}{2023}).
\newblock
\urldef\tempurl%
\url{https://www.aljazeera.com/news/2023/11/10/which-of-gazas-hospitals-is-israel-threatening}
\showURL{%
\tempurl}


\bibitem[{Al Jazeera}(2023d)]%
        {al_jazeera_we_2023_ed}
\bibfield{author}{\bibinfo{person}{{Al Jazeera}}.} \bibinfo{year}{2023}\natexlab{d}.
\newblock \showarticletitle{‘{We} can’t confuse {Hamas} with all {Palestinians}’: {Spain} says amid aid review}.
\newblock \bibinfo{journal}{\emph{Al Jazeera}} (\bibinfo{date}{Oct.} \bibinfo{year}{2023}).
\newblock
\urldef\tempurl%
\url{https://www.aljazeera.com/news/2023/10/10/we-cant-confuse-hamas-with-all-palestinians-spain-says-amid-aid-review}
\showURL{%
\tempurl}


\bibitem[{Al Jazeera}(2024)]%
        {al_jazeera_houthis_2024_ed}
\bibfield{author}{\bibinfo{person}{{Al Jazeera}}.} \bibinfo{year}{2024}\natexlab{}.
\newblock \showarticletitle{Houthis threaten more attacks against {Western} warships}.
\newblock \bibinfo{journal}{\emph{Al Jazeera}} (\bibinfo{date}{Jan.} \bibinfo{year}{2024}).
\newblock
\urldef\tempurl%
\url{https://www.aljazeera.com/news/2024/1/31/houthis-threaten-more-strikes-on-us-and-british-warships}
\showURL{%
\tempurl}


\bibitem[Ali and Hassan(2022)]%
        {ali_survey_2022}
\bibfield{author}{\bibinfo{person}{Mohammad Ali} {and} \bibinfo{person}{Naeemul Hassan}.} \bibinfo{year}{2022}\natexlab{}.
\newblock \showarticletitle{A {Survey} of {Computational} {Framing} {Analysis} {Approaches}}. In \bibinfo{booktitle}{\emph{Proceedings of the 2022 {Conference} on {Empirical} {Methods} in {Natural} {Language} {Processing}}}, \bibfield{editor}{\bibinfo{person}{Yoav Goldberg}, \bibinfo{person}{Zornitsa Kozareva}, {and} \bibinfo{person}{Yue Zhang}} (Eds.). \bibinfo{publisher}{Association for Computational Linguistics}, \bibinfo{address}{Abu Dhabi, United Arab Emirates}, \bibinfo{pages}{9335--9348}.
\newblock
\urldef\tempurl%
\url{https://doi.org/10.18653/v1/2022.emnlp-main.633}
\showDOI{\tempurl}


\bibitem[Ash et~al\mbox{.}(2024)]%
        {ash_relatio_2024}
\bibfield{author}{\bibinfo{person}{Elliott Ash}, \bibinfo{person}{Germain Gauthier}, {and} \bibinfo{person}{Philine Widmer}.} \bibinfo{year}{2024}\natexlab{}.
\newblock \showarticletitle{Relatio: {Text} {Semantics} {Capture} {Political} and {Economic} {Narratives}}.
\newblock \bibinfo{journal}{\emph{Political Analysis}} \bibinfo{volume}{32}, \bibinfo{number}{1} (\bibinfo{date}{Jan.} \bibinfo{year}{2024}), \bibinfo{pages}{115--132}.
\newblock
\showISSN{1047-1987, 1476-4989}
\urldef\tempurl%
\url{https://doi.org/10.1017/pan.2023.8}
\showDOI{\tempurl}


\bibitem[Aukes et~al\mbox{.}(2020)]%
        {aukes_narrative_2020}
\bibfield{author}{\bibinfo{person}{Ewert~Johannes Aukes}, \bibinfo{person}{Lotte~E. Bontje}, {and} \bibinfo{person}{Jill~H. Slinger}.} \bibinfo{year}{2020}\natexlab{}.
\newblock \showarticletitle{Narrative and {Frame} {Analysis}: {Disentangling} and {Refining} {Two} {Close} {Relatives} by {Means} of a {Large} {Infrastructural} {Technology} {Case}}.
\newblock \bibinfo{journal}{\emph{Forum Qualitative Sozialforschung / Forum: Qualitative Social Research}} \bibinfo{volume}{21}, \bibinfo{number}{2} (\bibinfo{date}{May} \bibinfo{year}{2020}).
\newblock
\showISSN{1438-5627}
\urldef\tempurl%
\url{https://doi.org/10.17169/fqs-21.2.3422}
\showDOI{\tempurl}


\bibitem[Baier(2023)]%
        {baier_narratives_2023}
\bibfield{author}{\bibinfo{person}{Christian Baier}.} \bibinfo{year}{2023}\natexlab{}.
\newblock \showarticletitle{Narratives of {Post}-{Truth}: {Lyotard} and the {Epistemic} {Fragmentation} of {Society}}.
\newblock \bibinfo{journal}{\emph{Theory, Culture \& Society}} \bibinfo{volume}{41}, \bibinfo{number}{1} (\bibinfo{date}{April} \bibinfo{year}{2023}), \bibinfo{pages}{1--16}.
\newblock
\showISSN{0263-2764}
\urldef\tempurl%
\url{https://doi.org/10.1177/02632764231162027}
\showDOI{\tempurl}


\bibitem[Bandeli et~al\mbox{.}(2020)]%
        {bandeli_framework_2020}
\bibfield{author}{\bibinfo{person}{Kiran~Kumar Bandeli}, \bibinfo{person}{Muhammad Hussain}, {and} \bibinfo{person}{Nitin Agarwal}.} \bibinfo{year}{2020}\natexlab{}.
\newblock \bibinfo{booktitle}{\emph{A {Framework} towards {Computational} {Narrative} {Analysis} on {Blogs}}}.
\newblock


\bibitem[Bar-Tal et~al\mbox{.}(2014)]%
        {bar-tal_sociopsychological_2014}
\bibfield{author}{\bibinfo{person}{Daniel Bar-Tal}, \bibinfo{person}{Neta Oren}, {and} \bibinfo{person}{Rafi Nets-Zehngut}.} \bibinfo{year}{2014}\natexlab{}.
\newblock \showarticletitle{Sociopsychological analysis of conflict-supporting narratives: {A} general framework}.
\newblock \bibinfo{journal}{\emph{Journal of Peace Research}} \bibinfo{volume}{51}, \bibinfo{number}{5} (\bibinfo{date}{Sept.} \bibinfo{year}{2014}), \bibinfo{pages}{662--675}.
\newblock
\showISSN{0022-3433}
\urldef\tempurl%
\url{https://doi.org/10.1177/0022343314533984}
\showDOI{\tempurl}
\newblock
\shownote{Publisher: SAGE Publications Ltd}.


\bibitem[BehnamGhader et~al\mbox{.}(2024)]%
        {behnamghader_llm2vec_2024}
\bibfield{author}{\bibinfo{person}{Parishad BehnamGhader}, \bibinfo{person}{Vaibhav Adlakha}, \bibinfo{person}{Marius Mosbach}, \bibinfo{person}{Dzmitry Bahdanau}, \bibinfo{person}{Nicolas Chapados}, {and} \bibinfo{person}{Siva Reddy}.} \bibinfo{year}{2024}\natexlab{}.
\newblock \bibinfo{title}{{LLM2Vec}: {Large} {Language} {Models} {Are} {Secretly} {Powerful} {Text} {Encoders}}.
\newblock
\newblock
\urldef\tempurl%
\url{https://doi.org/10.48550/arXiv.2404.05961}
\showDOI{\tempurl}
\newblock
\shownote{arXiv:2404.05961 [cs]}.


\bibitem[Blei et~al\mbox{.}(2003)]%
        {blei_latent_2003}
\bibfield{author}{\bibinfo{person}{David~M. Blei}, \bibinfo{person}{Andrew~Y. Ng}, {and} \bibinfo{person}{Michael~I. Jordan}.} \bibinfo{year}{2003}\natexlab{}.
\newblock \showarticletitle{Latent dirichlet allocation}.
\newblock \bibinfo{journal}{\emph{The Journal of Machine Learning Research}}  \bibinfo{volume}{3} (\bibinfo{year}{2003}), \bibinfo{pages}{993--1022}.
\newblock
\showISSN{1532-4435}


\bibitem[Bliuc and Chidley(2022)]%
        {bliuc_cooperation_2022}
\bibfield{author}{\bibinfo{person}{Ana-Maria Bliuc} {and} \bibinfo{person}{Alexander Chidley}.} \bibinfo{year}{2022}\natexlab{}.
\newblock \showarticletitle{From cooperation to conflict: {The} role of collective narratives in shaping group behaviour}.
\newblock \bibinfo{journal}{\emph{Social and Personality Psychology Compass}} \bibinfo{volume}{16}, \bibinfo{number}{7} (\bibinfo{year}{2022}), \bibinfo{pages}{e12670}.
\newblock
\showISSN{1751-9004}
\urldef\tempurl%
\url{https://doi.org/10.1111/spc3.12670}
\showDOI{\tempurl}


\bibitem[Bushell et~al\mbox{.}(2017)]%
        {bushell_strategic_2017}
\bibfield{author}{\bibinfo{person}{Simon Bushell}, \bibinfo{person}{Géraldine~Satre Buisson}, \bibinfo{person}{Mark Workman}, {and} \bibinfo{person}{Thomas Colley}.} \bibinfo{year}{2017}\natexlab{}.
\newblock \showarticletitle{Strategic narratives in climate change: {Towards} a unifying narrative to address the action gap on climate change}.
\newblock \bibinfo{journal}{\emph{Energy Research \& Social Science}}  \bibinfo{volume}{28} (\bibinfo{date}{June} \bibinfo{year}{2017}), \bibinfo{pages}{39--49}.
\newblock
\showISSN{2214-6296}
\urldef\tempurl%
\url{https://doi.org/10.1016/j.erss.2017.04.001}
\showDOI{\tempurl}


\bibitem[Chambers and Jurafsky(2008)]%
        {chambers_unsupervised_2008}
\bibfield{author}{\bibinfo{person}{Nathanael Chambers} {and} \bibinfo{person}{Dan Jurafsky}.} \bibinfo{year}{2008}\natexlab{}.
\newblock \showarticletitle{Unsupervised {Learning} of {Narrative} {Event} {Chains}}. In \bibinfo{booktitle}{\emph{Proceedings of {ACL}-08: {HLT}}}, \bibfield{editor}{\bibinfo{person}{Johanna~D. Moore}, \bibinfo{person}{Simone Teufel}, \bibinfo{person}{James Allan}, {and} \bibinfo{person}{Sadaoki Furui}} (Eds.). \bibinfo{publisher}{Association for Computational Linguistics}, \bibinfo{address}{Columbus, Ohio}, \bibinfo{pages}{789--797}.
\newblock
\urldef\tempurl%
\url{https://aclanthology.org/P08-1090}
\showURL{%
\tempurl}


\bibitem[Chaturvedi et~al\mbox{.}(2017)]%
        {chaturvedi_unsupervised_2017}
\bibfield{author}{\bibinfo{person}{Snigdha Chaturvedi}, \bibinfo{person}{Mohit Iyyer}, {and} \bibinfo{person}{Hal~Daume Iii}.} \bibinfo{year}{2017}\natexlab{}.
\newblock \showarticletitle{Unsupervised {Learning} of {Evolving} {Relationships} {Between} {Literary} {Characters}}.
\newblock \bibinfo{journal}{\emph{Proceedings of the AAAI Conference on Artificial Intelligence}} \bibinfo{volume}{31}, \bibinfo{number}{1} (\bibinfo{date}{Feb.} \bibinfo{year}{2017}).
\newblock
\showISSN{2374-3468}
\urldef\tempurl%
\url{https://doi.org/10.1609/aaai.v31i1.10982}
\showDOI{\tempurl}


\bibitem[Coan et~al\mbox{.}(2021)]%
        {coan_computer-assisted_2021}
\bibfield{author}{\bibinfo{person}{Travis~G. Coan}, \bibinfo{person}{Constantine Boussalis}, \bibinfo{person}{John Cook}, {and} \bibinfo{person}{Mirjam~O. Nanko}.} \bibinfo{year}{2021}\natexlab{}.
\newblock \showarticletitle{Computer-assisted classification of contrarian claims about climate change}.
\newblock \bibinfo{journal}{\emph{Scientific Reports}} \bibinfo{volume}{11}, \bibinfo{number}{1} (\bibinfo{date}{Nov.} \bibinfo{year}{2021}), \bibinfo{pages}{22320}.
\newblock
\showISSN{2045-2322}
\urldef\tempurl%
\url{https://doi.org/10.1038/s41598-021-01714-4}
\showDOI{\tempurl}


\bibitem[de~Saint~Laurent et~al\mbox{.}(2021)]%
        {de_saint_laurent_internet_2021}
\bibfield{author}{\bibinfo{person}{Constance de Saint~Laurent}, \bibinfo{person}{Vlad~P. Glăveanu}, {and} \bibinfo{person}{Ioana Literat}.} \bibinfo{year}{2021}\natexlab{}.
\newblock \showarticletitle{Internet {Memes} as {Partial} {Stories}: {Identifying} {Political} {Narratives} in {Coronavirus} {Memes}}.
\newblock \bibinfo{journal}{\emph{Social Media + Society}} \bibinfo{volume}{7}, \bibinfo{number}{1} (\bibinfo{date}{Jan.} \bibinfo{year}{2021}).
\newblock
\showISSN{2056-3051}
\urldef\tempurl%
\url{https://doi.org/10.1177/2056305121988932}
\showDOI{\tempurl}


\bibitem[Devlin et~al\mbox{.}(2019)]%
        {devlin_bert_2019}
\bibfield{author}{\bibinfo{person}{Jacob Devlin}, \bibinfo{person}{Ming-Wei Chang}, \bibinfo{person}{Kenton Lee}, {and} \bibinfo{person}{Kristina Toutanova}.} \bibinfo{year}{2019}\natexlab{}.
\newblock \bibinfo{title}{{BERT}: {Pre}-training of {Deep} {Bidirectional} {Transformers} for {Language} {Understanding}}.
\newblock
\newblock
\urldef\tempurl%
\url{https://doi.org/10.48550/arXiv.1810.04805}
\showDOI{\tempurl}
\newblock
\shownote{arXiv:1810.04805 [cs]}.


\bibitem[Dubey et~al\mbox{.}(2024)]%
        {dubey_llama_2024_ed}
\bibfield{author}{\bibinfo{person}{Abhimanyu Dubey}, \bibinfo{person}{Abhinav Jauhri}, \bibinfo{person}{Abhinav Pandey}, \bibinfo{person}{Abhishek Kadian}, \bibinfo{person}{Ahmad Al-Dahle}, \bibinfo{person}{Aiesha Letman}, \bibinfo{person}{Akhil Mathur}, \bibinfo{person}{Alan Schelten}, \bibinfo{person}{Amy Yang}, \bibinfo{person}{Angela Fan}, \bibinfo{person}{Anirudh Goyal}, \bibinfo{person}{Anthony Hartshorn}, \bibinfo{person}{Aobo Yang}, \bibinfo{person}{Archi Mitra}, \bibinfo{person}{Archie Sravankumar}, \bibinfo{person}{Artem Korenev}, \bibinfo{person}{Arthur Hinsvark}, \bibinfo{person}{Arun Rao}, \bibinfo{person}{Aston {Zhang, ...}}, {and} \bibinfo{person}{Zhiwei Zhao}.} \bibinfo{year}{2024}\natexlab{}.
\newblock \bibinfo{title}{The {Llama} 3 {Herd} of {Models}}.
\newblock
\newblock
\urldef\tempurl%
\url{https://doi.org/10.48550/arXiv.2407.21783}
\showDOI{\tempurl}
\newblock
\shownote{arXiv:2407.21783 [cs]}.


\bibitem[Entman(1993)]%
        {entman_framing_1993}
\bibfield{author}{\bibinfo{person}{Robert~M. Entman}.} \bibinfo{year}{1993}\natexlab{}.
\newblock \showarticletitle{Framing: {Toward} {Clarification} of a {Fractured} {Paradigm}}.
\newblock \bibinfo{journal}{\emph{Journal of Communication}} \bibinfo{volume}{43}, \bibinfo{number}{4} (\bibinfo{year}{1993}), \bibinfo{pages}{51--58}.
\newblock
\showISSN{1460-2466}
\urldef\tempurl%
\url{https://doi.org/10.1111/j.1460-2466.1993.tb01304.x}
\showDOI{\tempurl}


\bibitem[Fahmy and Al~Emad(2011)]%
        {fahmy_-jazeera_2011}
\bibfield{author}{\bibinfo{person}{Shahira~S Fahmy} {and} \bibinfo{person}{Mohammed Al~Emad}.} \bibinfo{year}{2011}\natexlab{}.
\newblock \showarticletitle{Al-{Jazeera} vs {Al}-{Jazeera}: {A} comparison of the network’s {English} and {Arabic} online coverage of the {US}/{Al} {Qaeda} conflict}.
\newblock \bibinfo{journal}{\emph{International Communication Gazette}} \bibinfo{volume}{73}, \bibinfo{number}{3} (\bibinfo{date}{April} \bibinfo{year}{2011}), \bibinfo{pages}{216--232}.
\newblock
\showISSN{1748-0485}
\urldef\tempurl%
\url{https://doi.org/10.1177/1748048510393656}
\showDOI{\tempurl}
\newblock
\shownote{Publisher: SAGE Publications Ltd}.


\bibitem[George(2023)]%
        {george_amid_2023}
\bibfield{author}{\bibinfo{person}{Susannah George}.} \bibinfo{year}{2023}\natexlab{}.
\newblock \showarticletitle{Amid so much death, {Israelis} struggle to lay their loved ones to rest}.
\newblock \bibinfo{journal}{\emph{Washington Post}} (\bibinfo{date}{Oct.} \bibinfo{year}{2023}).
\newblock
\showISSN{0190-8286}
\urldef\tempurl%
\url{https://www.washingtonpost.com/world/2023/10/09/israel-hamas-attack-deal-toll/}
\showURL{%
\tempurl}


\bibitem[Gervás and Méndez(2024)]%
        {gervas_tagging_2024}
\bibfield{author}{\bibinfo{person}{Pablo Gervás} {and} \bibinfo{person}{Gonzalo Méndez}.} \bibinfo{year}{2024}\natexlab{}.
\newblock \showarticletitle{Tagging {Narrative} with {Propp}’s {Character} {Functions} {Using} {Large} {Language} {Models}}. In \bibinfo{booktitle}{\emph{Proceedings of the {Text2Story}’24 {Workshop}}}. \bibinfo{publisher}{CEUR-WS.org}, \bibinfo{address}{Glasgow}, \bibinfo{pages}{137--148}.
\newblock
\urldef\tempurl%
\url{https://ceur-ws.org/Vol-3671/paper12.pdf}
\showURL{%
\tempurl}


\bibitem[Gildea and Jurafsky(2002)]%
        {gildea_automatic_2002}
\bibfield{author}{\bibinfo{person}{Daniel Gildea} {and} \bibinfo{person}{Daniel Jurafsky}.} \bibinfo{year}{2002}\natexlab{}.
\newblock \showarticletitle{Automatic {Labeling} of {Semantic} {Roles}}.
\newblock \bibinfo{journal}{\emph{Computational Linguistics}} \bibinfo{volume}{28}, \bibinfo{number}{3} (\bibinfo{date}{Sept.} \bibinfo{year}{2002}), \bibinfo{pages}{245--288}.
\newblock
\showISSN{0891-2017}
\urldef\tempurl%
\url{https://doi.org/10.1162/089120102760275983}
\showDOI{\tempurl}


\bibitem[Goodwin(2024)]%
        {goodwin_senate_2024}
\bibfield{author}{\bibinfo{person}{Liz Goodwin}.} \bibinfo{year}{2024}\natexlab{}.
\newblock \showarticletitle{Senate votes to advance {Ukraine}-{Israel} package after border deal fails}.
\newblock \bibinfo{journal}{\emph{Washington Post}} (\bibinfo{date}{Feb.} \bibinfo{year}{2024}).
\newblock
\showISSN{0190-8286}
\urldef\tempurl%
\url{https://www.washingtonpost.com/politics/2024/02/08/senate-ukraine-israel-aid-border-deal/}
\showURL{%
\tempurl}


\bibitem[Greimas(1984)]%
        {greimas_structural_1984}
\bibfield{author}{\bibinfo{person}{Algirdas~Julien Greimas}.} \bibinfo{year}{1984}\natexlab{}.
\newblock \bibinfo{booktitle}{\emph{Structural semantics: an attempt at a method}}.
\newblock \bibinfo{publisher}{University of Nebraska Press}, \bibinfo{address}{Lincoln}.
\newblock
\showISBNx{978-0-8032-2112-3}


\bibitem[Grootendorst(2022)]%
        {grootendorst_bertopic_2022}
\bibfield{author}{\bibinfo{person}{Maarten Grootendorst}.} \bibinfo{year}{2022}\natexlab{}.
\newblock \bibinfo{title}{{BERTopic}: {Neural} topic modeling with a class-based {TF}-{IDF} procedure}.
\newblock
\newblock
\urldef\tempurl%
\url{https://doi.org/10.48550/arXiv.2203.05794}
\showDOI{\tempurl}
\newblock
\shownote{arXiv:2203.05794 [cs]}.


\bibitem[Herman(2001)]%
        {herman_narrative_2001}
\bibfield{author}{\bibinfo{person}{David Herman}.} \bibinfo{year}{2001}\natexlab{}.
\newblock \showarticletitle{Narrative {Theory} and the {Cognitive} {Sciences}}.
\newblock \bibinfo{journal}{\emph{Narrative Inquiry}} \bibinfo{volume}{11}, \bibinfo{number}{1} (\bibinfo{date}{Jan.} \bibinfo{year}{2001}), \bibinfo{pages}{1--34}.
\newblock
\showISSN{1387-6740, 1569-9935}
\urldef\tempurl%
\url{https://doi.org/10.1075/ni.11.1.01her}
\showDOI{\tempurl}


\bibitem[Ignatius(2023)]%
        {ignatius_opinion_2023}
\bibfield{author}{\bibinfo{person}{David Ignatius}.} \bibinfo{year}{2023}\natexlab{}.
\newblock \showarticletitle{Opinion {\textbar} {A} fraught battlespace awaits {Israel} after the pause}.
\newblock \bibinfo{journal}{\emph{Washington Post}} (\bibinfo{date}{Nov.} \bibinfo{year}{2023}).
\newblock
\showISSN{0190-8286}
\urldef\tempurl%
\url{https://www.washingtonpost.com/opinions/2023/11/25/israel-gaza-war-hamas-ceasefire-next-phase/}
\showURL{%
\tempurl}


\bibitem[Kusovac(2023)]%
        {kusovac_how_2023}
\bibfield{author}{\bibinfo{person}{Zoran Kusovac}.} \bibinfo{year}{2023}\natexlab{}.
\newblock \showarticletitle{How {Israel} will stage its land incursion into {Gaza}}.
\newblock \bibinfo{journal}{\emph{Al Jazeera}} (\bibinfo{date}{Oct.} \bibinfo{year}{2023}).
\newblock
\urldef\tempurl%
\url{https://www.aljazeera.com/news/2023/10/20/how-israel-will-stage-its-land-incursion-into-gaza}
\showURL{%
\tempurl}


\bibitem[Labatut and Bost(2020)]%
        {labatut_extraction_2020}
\bibfield{author}{\bibinfo{person}{Vincent Labatut} {and} \bibinfo{person}{Xavier Bost}.} \bibinfo{year}{2020}\natexlab{}.
\newblock \showarticletitle{Extraction and {Analysis} of {Fictional} {Character} {Networks}: {A} {Survey}}.
\newblock \bibinfo{journal}{\emph{Comput. Surveys}} \bibinfo{volume}{52}, \bibinfo{number}{5} (\bibinfo{date}{Sept.} \bibinfo{year}{2020}), \bibinfo{pages}{1--40}.
\newblock
\showISSN{0360-0300, 1557-7341}
\urldef\tempurl%
\url{https://doi.org/10.1145/3344548}
\showDOI{\tempurl}


\bibitem[Maines(1993)]%
        {maines_narratives_1993}
\bibfield{author}{\bibinfo{person}{David~R. Maines}.} \bibinfo{year}{1993}\natexlab{}.
\newblock \showarticletitle{Narrative's {Moment} and {Sociology}'s {Phenomena}: {Toward} a {Narrative} {Sociology}}.
\newblock \bibinfo{journal}{\emph{The Sociological Quarterly}} \bibinfo{volume}{34}, \bibinfo{number}{1} (\bibinfo{year}{1993}), \bibinfo{pages}{17--38}.
\newblock
\showISSN{0038-0253}
\urldef\tempurl%
\url{https://www.jstor.org/stable/4121556}
\showURL{%
\tempurl}


\bibitem[McInnes et~al\mbox{.}(2020)]%
        {mcinnes_umap_2020}
\bibfield{author}{\bibinfo{person}{Leland McInnes}, \bibinfo{person}{John Healy}, {and} \bibinfo{person}{James Melville}.} \bibinfo{year}{2020}\natexlab{}.
\newblock \bibinfo{title}{{UMAP}: {Uniform} {Manifold} {Approximation} and {Projection} for {Dimension} {Reduction}}.
\newblock
\newblock
\urldef\tempurl%
\url{https://doi.org/10.48550/arXiv.1802.03426}
\showDOI{\tempurl}
\newblock
\shownote{arXiv:1802.03426 [cs, stat]}.


\bibitem[Mohr et~al\mbox{.}(2013)]%
        {mohr_graphing_2013}
\bibfield{author}{\bibinfo{person}{John~W. Mohr}, \bibinfo{person}{Robin Wagner-Pacifici}, \bibinfo{person}{Ronald~L. Breiger}, {and} \bibinfo{person}{Petko Bogdanov}.} \bibinfo{year}{2013}\natexlab{}.
\newblock \showarticletitle{Graphing the grammar of motives in {National} {Security} {Strategies}: {Cultural} interpretation, automated text analysis and the drama of global politics}.
\newblock \bibinfo{journal}{\emph{Poetics}} \bibinfo{volume}{41}, \bibinfo{number}{6} (\bibinfo{date}{Dec.} \bibinfo{year}{2013}), \bibinfo{pages}{670--700}.
\newblock
\showISSN{0304-422X}
\urldef\tempurl%
\url{https://doi.org/10.1016/j.poetic.2013.08.003}
\showDOI{\tempurl}


\bibitem[MTEB(2024)]%
        {mteb_mteb_2024}
\bibfield{author}{\bibinfo{person}{MTEB}.} \bibinfo{year}{2024}\natexlab{}.
\newblock \bibinfo{title}{{MTEB} {Leaderboard}}.
\newblock
\newblock
\urldef\tempurl%
\url{https://huggingface.co/spaces/mteb/leaderboard}
\showURL{%
\tempurl}


\bibitem[Muennighoff et~al\mbox{.}(2023)]%
        {muennighoff_mteb_2023}
\bibfield{author}{\bibinfo{person}{Niklas Muennighoff}, \bibinfo{person}{Nouamane Tazi}, \bibinfo{person}{Loic Magne}, {and} \bibinfo{person}{Nils Reimers}.} \bibinfo{year}{2023}\natexlab{}.
\newblock \showarticletitle{{MTEB}: {Massive} {Text} {Embedding} {Benchmark}}. In \bibinfo{booktitle}{\emph{Proceedings of the 17th {Conference} of the {European} {Chapter} of the {Association} for {Computational} {Linguistics}}}, \bibfield{editor}{\bibinfo{person}{Andreas Vlachos} {and} \bibinfo{person}{Isabelle Augenstein}} (Eds.). \bibinfo{publisher}{Association for Computational Linguistics}, \bibinfo{address}{Dubrovnik, Croatia}, \bibinfo{pages}{2014--2037}.
\newblock
\urldef\tempurl%
\url{https://doi.org/10.18653/v1/2023.eacl-main.148}
\showDOI{\tempurl}


\bibitem[Parker et~al\mbox{.}(2024)]%
        {parker_what_2024}
\bibfield{author}{\bibinfo{person}{Claire Parker}, \bibinfo{person}{Annabelle Timsit}, {and} \bibinfo{person}{Adam Taylor}.} \bibinfo{year}{2024}\natexlab{}.
\newblock \showarticletitle{What qualifies as genocide? {Breaking} down the {ICJ} case against {Israel}.}
\newblock \bibinfo{journal}{\emph{Washington Post}} (\bibinfo{date}{Jan.} \bibinfo{year}{2024}).
\newblock
\showISSN{0190-8286}
\urldef\tempurl%
\url{https://www.washingtonpost.com/world/2024/01/11/what-is-genocide-definition-courts/}
\showURL{%
\tempurl}


\bibitem[Piper et~al\mbox{.}(2021)]%
        {piper_narrative_2021}
\bibfield{author}{\bibinfo{person}{Andrew Piper}, \bibinfo{person}{Richard~Jean So}, {and} \bibinfo{person}{David Bamman}.} \bibinfo{year}{2021}\natexlab{}.
\newblock \showarticletitle{Narrative {Theory} for {Computational} {Narrative} {Understanding}}. In \bibinfo{booktitle}{\emph{Proceedings of the 2021 {Conference} on {Empirical} {Methods} in {Natural} {Language} {Processing}}}, \bibfield{editor}{\bibinfo{person}{Marie-Francine Moens}, \bibinfo{person}{Xuanjing Huang}, \bibinfo{person}{Lucia Specia}, {and} \bibinfo{person}{Scott Wen-tau Yih}} (Eds.). \bibinfo{publisher}{Association for Computational Linguistics}, \bibinfo{address}{Online and Punta Cana, Dominican Republic}, \bibinfo{pages}{298--311}.
\newblock
\urldef\tempurl%
\url{https://doi.org/10.18653/v1/2021.emnlp-main.26}
\showDOI{\tempurl}


\bibitem[Polletta and Callahan(2017)]%
        {polletta_deep_2017}
\bibfield{author}{\bibinfo{person}{Francesca Polletta} {and} \bibinfo{person}{Jessica Callahan}.} \bibinfo{year}{2017}\natexlab{}.
\newblock \showarticletitle{Deep stories, nostalgia narratives, and fake news: {Storytelling} in the {Trump} era}.
\newblock \bibinfo{journal}{\emph{American Journal of Cultural Sociology}} \bibinfo{volume}{5}, \bibinfo{number}{3} (\bibinfo{date}{Oct.} \bibinfo{year}{2017}), \bibinfo{pages}{392--408}.
\newblock
\showISSN{2049-7121}
\urldef\tempurl%
\url{https://doi.org/10.1057/s41290-017-0037-7}
\showDOI{\tempurl}


\bibitem[Propp(1968)]%
        {propp_morphology_1968}
\bibfield{author}{\bibinfo{person}{Vladimir Propp}.} \bibinfo{year}{1968}\natexlab{}.
\newblock \bibinfo{booktitle}{\emph{Morphology of the {Folktale}} (\bibinfo{edition}{2} ed.)}.
\newblock \bibinfo{publisher}{University of Texas Press}, \bibinfo{address}{Austin}.
\newblock
\showISBNx{978-0-292-78391-1}
\urldef\tempurl%
\url{https://doi.org/10.7560/783911}
\showDOI{\tempurl}


\bibitem[Ranade et~al\mbox{.}(2022)]%
        {ranade_computational_2022}
\bibfield{author}{\bibinfo{person}{Priyanka Ranade}, \bibinfo{person}{Sanorita Dey}, \bibinfo{person}{Anupam Joshi}, {and} \bibinfo{person}{Tim Finin}.} \bibinfo{year}{2022}\natexlab{}.
\newblock \showarticletitle{Computational {Understanding} of {Narratives}: {A} {Survey}}.
\newblock \bibinfo{journal}{\emph{IEEE Access}}  \bibinfo{volume}{10} (\bibinfo{year}{2022}), \bibinfo{pages}{101575--101594}.
\newblock
\showISSN{2169-3536}
\urldef\tempurl%
\url{https://doi.org/10.1109/ACCESS.2022.3205314}
\showDOI{\tempurl}


\bibitem[Reimers and Gurevych(2019)]%
        {reimers_sentence-bert_2019}
\bibfield{author}{\bibinfo{person}{Nils Reimers} {and} \bibinfo{person}{Iryna Gurevych}.} \bibinfo{year}{2019}\natexlab{}.
\newblock \showarticletitle{Sentence-{BERT}: {Sentence} {Embeddings} using {Siamese} {BERT}-{Networks}}. In \bibinfo{booktitle}{\emph{Proceedings of the 2019 {Conference} on {Empirical} {Methods} in {Natural} {Language} {Processing} and the 9th {International} {Joint} {Conference} on {Natural} {Language} {Processing} ({EMNLP}-{IJCNLP})}}, \bibfield{editor}{\bibinfo{person}{Kentaro Inui}, \bibinfo{person}{Jing Jiang}, \bibinfo{person}{Vincent Ng}, {and} \bibinfo{person}{Xiaojun Wan}} (Eds.). \bibinfo{publisher}{Association for Computational Linguistics}, \bibinfo{address}{Hong Kong, China}, \bibinfo{pages}{3982--3992}.
\newblock
\urldef\tempurl%
\url{https://doi.org/10.18653/v1/D19-1410}
\showDOI{\tempurl}


\bibitem[Reiter-Haas et~al\mbox{.}(2024)]%
        {reiter-haas_computational_2024}
\bibfield{author}{\bibinfo{person}{Markus Reiter-Haas}, \bibinfo{person}{Beate Klösch}, \bibinfo{person}{Markus Hadler}, {and} \bibinfo{person}{Elisabeth Lex}.} \bibinfo{year}{2024}\natexlab{}.
\newblock \showarticletitle{Computational {Narrative} {Framing}: {Towards} {Identifying} {Frames} through {Contrasting} the {Evolution} of {Narrations}}. In \bibinfo{booktitle}{\emph{Proceedings of the {Text2Story}'24 {Workshop}}}. \bibinfo{publisher}{CEUR-WS.org}, \bibinfo{address}{Glasgow}.
\newblock
\urldef\tempurl%
\url{https://ceur-ws.org/Vol-3671/paper11.pdf}
\showURL{%
\tempurl}


\bibitem[Roberts et~al\mbox{.}(2021)]%
        {roberts_media_2021}
\bibfield{author}{\bibinfo{person}{Hal Roberts}, \bibinfo{person}{Rahul Bhargava}, \bibinfo{person}{Linas Valiukas}, \bibinfo{person}{Dennis Jen}, \bibinfo{person}{Momin~M. Malik}, \bibinfo{person}{Cindy~Sherman Bishop}, \bibinfo{person}{Emily~B. Ndulue}, \bibinfo{person}{Aashka Dave}, \bibinfo{person}{Justin Clark}, \bibinfo{person}{Bruce Etling}, \bibinfo{person}{Robert Faris}, \bibinfo{person}{Anushka Shah}, \bibinfo{person}{Jasmin Rubinovitz}, \bibinfo{person}{Alexis Hope}, \bibinfo{person}{Catherine D'Ignazio}, \bibinfo{person}{Fernando Bermejo}, \bibinfo{person}{Yochai Benkler}, {and} \bibinfo{person}{Ethan Zuckerman}.} \bibinfo{year}{2021}\natexlab{}.
\newblock \showarticletitle{Media {Cloud}: {Massive} {Open} {Source} {Collection} of {Global} {News} on the {Open} {Web}}.
\newblock \bibinfo{journal}{\emph{Proceedings of the International AAAI Conference on Web and Social Media}}  \bibinfo{volume}{15} (\bibinfo{date}{May} \bibinfo{year}{2021}), \bibinfo{pages}{1034--1045}.
\newblock
\showISSN{2334-0770, 2162-3449}
\urldef\tempurl%
\url{https://doi.org/10.1609/icwsm.v15i1.18127}
\showDOI{\tempurl}


\bibitem[Rubin(2023)]%
        {rubin_opinion_2023}
\bibfield{author}{\bibinfo{person}{Jennifer Rubin}.} \bibinfo{year}{2023}\natexlab{}.
\newblock \showarticletitle{Opinion {\textbar} {Fight} moral nihilism and relativism about {Israel}. {Facts} matter.}
\newblock \bibinfo{journal}{\emph{Washington Post}} (\bibinfo{date}{Nov.} \bibinfo{year}{2023}).
\newblock
\showISSN{0190-8286}
\urldef\tempurl%
\url{https://www.washingtonpost.com/opinions/2023/11/03/israel-newsletter-moral-idiocy/}
\showURL{%
\tempurl}


\bibitem[Rubin et~al\mbox{.}(2023)]%
        {rubin_what_2023}
\bibfield{author}{\bibinfo{person}{Shira Rubin}, \bibinfo{person}{Miriam Berger}, {and} \bibinfo{person}{Adam Taylor}.} \bibinfo{year}{2023}\natexlab{}.
\newblock \showarticletitle{What to know about {Israel}’s protests and judicial overhaul plan}.
\newblock \bibinfo{journal}{\emph{Washington Post}} (\bibinfo{date}{March} \bibinfo{year}{2023}).
\newblock
\showISSN{0190-8286}
\urldef\tempurl%
\url{https://www.washingtonpost.com/world/2023/03/27/israel-protests-judicial-reform/}
\showURL{%
\tempurl}


\bibitem[Santana et~al\mbox{.}(2023)]%
        {santana_survey_2023}
\bibfield{author}{\bibinfo{person}{Brenda Santana}, \bibinfo{person}{Ricardo Campos}, \bibinfo{person}{Evelin Amorim}, \bibinfo{person}{Alípio Jorge}, \bibinfo{person}{Purificação Silvano}, {and} \bibinfo{person}{Sérgio Nunes}.} \bibinfo{year}{2023}\natexlab{}.
\newblock \showarticletitle{A survey on narrative extraction from textual data}.
\newblock \bibinfo{journal}{\emph{Artificial Intelligence Review}} \bibinfo{volume}{56}, \bibinfo{number}{8} (\bibinfo{date}{Aug.} \bibinfo{year}{2023}), \bibinfo{pages}{8393--8435}.
\newblock
\showISSN{1573-7462}
\urldef\tempurl%
\url{https://doi.org/10.1007/s10462-022-10338-7}
\showDOI{\tempurl}


\bibitem[Sarbin(1986)]%
        {sarbin_narrative_1986}
\bibfield{author}{\bibinfo{person}{Theodore~R. Sarbin}.} \bibinfo{year}{1986}\natexlab{}.
\newblock \bibinfo{booktitle}{\emph{Narrative {Psychology}: {The} {Storied} {Nature} of {Human} {Conduct}}}.
\newblock \bibinfo{publisher}{Bloomsbury Academic}, \bibinfo{address}{London}.
\newblock
\showISBNx{978-0-275-92103-3}


\bibitem[Shiller(2019)]%
        {shiller_narrative_2019}
\bibfield{author}{\bibinfo{person}{Robert~J. Shiller}.} \bibinfo{year}{2019}\natexlab{}.
\newblock \bibinfo{booktitle}{\emph{Narrative {Economics}: {How} {Stories} {Go} {Viral} and {Drive} {Major} {Economic} {Events}}}.
\newblock \bibinfo{publisher}{Princeton University Press}, \bibinfo{address}{Princeton}.
\newblock
\showISBNx{978-0-691-18229-2}


\bibitem[Stammbach et~al\mbox{.}(2022)]%
        {stammbach_heroes_2022}
\bibfield{author}{\bibinfo{person}{Dominik Stammbach}, \bibinfo{person}{Maria Antoniak}, {and} \bibinfo{person}{Elliott Ash}.} \bibinfo{year}{2022}\natexlab{}.
\newblock \bibinfo{title}{Heroes, {Villains}, and {Victims}, and {GPT}-3: {Automated} {Extraction} of {Character} {Roles} {Without} {Training} {Data}}.
\newblock
\newblock
\urldef\tempurl%
\url{https://doi.org/10.48550/arXiv.2205.07557}
\showDOI{\tempurl}


\bibitem[Tangherlini et~al\mbox{.}(2020)]%
        {tangherlini_automated_2020}
\bibfield{author}{\bibinfo{person}{Timothy~R. Tangherlini}, \bibinfo{person}{Shadi Shahsavari}, \bibinfo{person}{Behnam Shahbazi}, \bibinfo{person}{Ehsan Ebrahimzadeh}, {and} \bibinfo{person}{Vwani Roychowdhury}.} \bibinfo{year}{2020}\natexlab{}.
\newblock \showarticletitle{An automated pipeline for the discovery of conspiracy and conspiracy theory narrative frameworks: {Bridgegate}, {Pizzagate} and storytelling on the web}.
\newblock \bibinfo{journal}{\emph{PLOS ONE}} \bibinfo{volume}{15}, \bibinfo{number}{6} (\bibinfo{date}{June} \bibinfo{year}{2020}), \bibinfo{pages}{e0233879}.
\newblock
\showISSN{1932-6203}
\urldef\tempurl%
\url{https://doi.org/10.1371/journal.pone.0233879}
\showDOI{\tempurl}


\bibitem[Tharoor(2023)]%
        {tharoor_analysis_2023}
\bibfield{author}{\bibinfo{person}{Ishaan Tharoor}.} \bibinfo{year}{2023}\natexlab{}.
\newblock \showarticletitle{Analysis {\textbar} {Between} {Israel} and {Gaza}, a deep history of trauma and violence}.
\newblock \bibinfo{journal}{\emph{Washington Post}} (\bibinfo{date}{Oct.} \bibinfo{year}{2023}).
\newblock
\showISSN{0190-8286}
\urldef\tempurl%
\url{https://www.washingtonpost.com/world/2023/10/10/gaza-strip-israel-history-violence-hamas-occupation/}
\showURL{%
\tempurl}


\bibitem[Wang et~al\mbox{.}(2024)]%
        {wang_improving_2024}
\bibfield{author}{\bibinfo{person}{Liang Wang}, \bibinfo{person}{Nan Yang}, \bibinfo{person}{Xiaolong Huang}, \bibinfo{person}{Linjun Yang}, \bibinfo{person}{Rangan Majumder}, {and} \bibinfo{person}{Furu Wei}.} \bibinfo{year}{2024}\natexlab{}.
\newblock \bibinfo{title}{Improving {Text} {Embeddings} with {Large} {Language} {Models}}.
\newblock
\newblock
\urldef\tempurl%
\url{https://doi.org/10.48550/arXiv.2401.00368}
\showDOI{\tempurl}
\newblock
\shownote{arXiv:2401.00368 [cs]}.


\bibitem[Wang et~al\mbox{.}(2023)]%
        {wang_gpt-ner_2023}
\bibfield{author}{\bibinfo{person}{Shuhe Wang}, \bibinfo{person}{Xiaofei Sun}, \bibinfo{person}{Xiaoya Li}, \bibinfo{person}{Rongbin Ouyang}, \bibinfo{person}{Fei Wu}, \bibinfo{person}{Tianwei Zhang}, \bibinfo{person}{Jiwei Li}, {and} \bibinfo{person}{Guoyin Wang}.} \bibinfo{year}{2023}\natexlab{}.
\newblock \bibinfo{title}{{GPT}-{NER}: {Named} {Entity} {Recognition} via {Large} {Language} {Models}}.
\newblock
\newblock
\urldef\tempurl%
\url{https://doi.org/10.48550/arXiv.2304.10428}
\showDOI{\tempurl}
\newblock
\shownote{arXiv:2304.10428 [cs]}.


\bibitem[Wang et~al\mbox{.}(2020)]%
        {wang_minilm_2020}
\bibfield{author}{\bibinfo{person}{Wenhui Wang}, \bibinfo{person}{Furu Wei}, \bibinfo{person}{Li Dong}, \bibinfo{person}{Hangbo Bao}, \bibinfo{person}{Nan Yang}, {and} \bibinfo{person}{Ming Zhou}.} \bibinfo{year}{2020}\natexlab{}.
\newblock \bibinfo{title}{{MiniLM}: {Deep} {Self}-{Attention} {Distillation} for {Task}-{Agnostic} {Compression} of {Pre}-{Trained} {Transformers}}.
\newblock
\newblock
\urldef\tempurl%
\url{https://doi.org/10.48550/arXiv.2002.10957}
\showDOI{\tempurl}
\newblock
\shownote{arXiv:2002.10957 [cs]}.


\bibitem[Zhang et~al\mbox{.}(2024)]%
        {zhang_evaluating_2024}
\bibfield{author}{\bibinfo{person}{Gaifan Zhang}, \bibinfo{person}{Yi Zhou}, {and} \bibinfo{person}{Danushka Bollegala}.} \bibinfo{year}{2024}\natexlab{}.
\newblock \showarticletitle{Evaluating {Unsupervised} {Dimensionality} {Reduction} {Methods} for {Pretrained} {Sentence} {Embeddings}}. In \bibinfo{booktitle}{\emph{Proceedings of the 2024 {Joint} {International} {Conference} on {Computational} {Linguistics}, {Language} {Resources} and {Evaluation} ({LREC}-{COLING} 2024)}}, \bibfield{editor}{\bibinfo{person}{Nicoletta Calzolari}, \bibinfo{person}{Min-Yen Kan}, \bibinfo{person}{Veronique Hoste}, \bibinfo{person}{Alessandro Lenci}, \bibinfo{person}{Sakriani Sakti}, {and} \bibinfo{person}{Nianwen Xue}} (Eds.). \bibinfo{publisher}{ELRA and ICCL}, \bibinfo{address}{Torino, Italia}, \bibinfo{pages}{6530--6543}.
\newblock
\urldef\tempurl%
\url{https://aclanthology.org/2024.lrec-main.579}
\showURL{%
\tempurl}


\end{thebibliography}
% \bibliographystyle{acl_natbib}

\clearpage

\appendix
\renewcommand\thefigure{\thesection.\arabic{figure}}
\setcounter{figure}{0}

\renewcommand\thetable{\thesection.\arabic{table}}
\setcounter{table}{0}

% \section{Appendix}\label{sec:appendix}

\subsection{Dimension reduction}\label{sec:ab_dim_reduction}
As described in section \ref{sec:dim_reduction}, reducing the dimension of individual actant embeddings is supposed to reduce variance between highly similar actants, and make it easier for the clustering algorithm to identify dense areas. In this section, we explore, how dimension reduction impacts the semantic similarity behaviour of the embeddings. It is important to note, that the distribution of actants within the corpus impacts the behaviour of the dimensionality reduction technique. Hence, we apply the method to the non-unique set of all actants.

\subsubsection*{Average similarity}
\begin{figure}
    \centering
    \includegraphics[width=.5\textwidth]{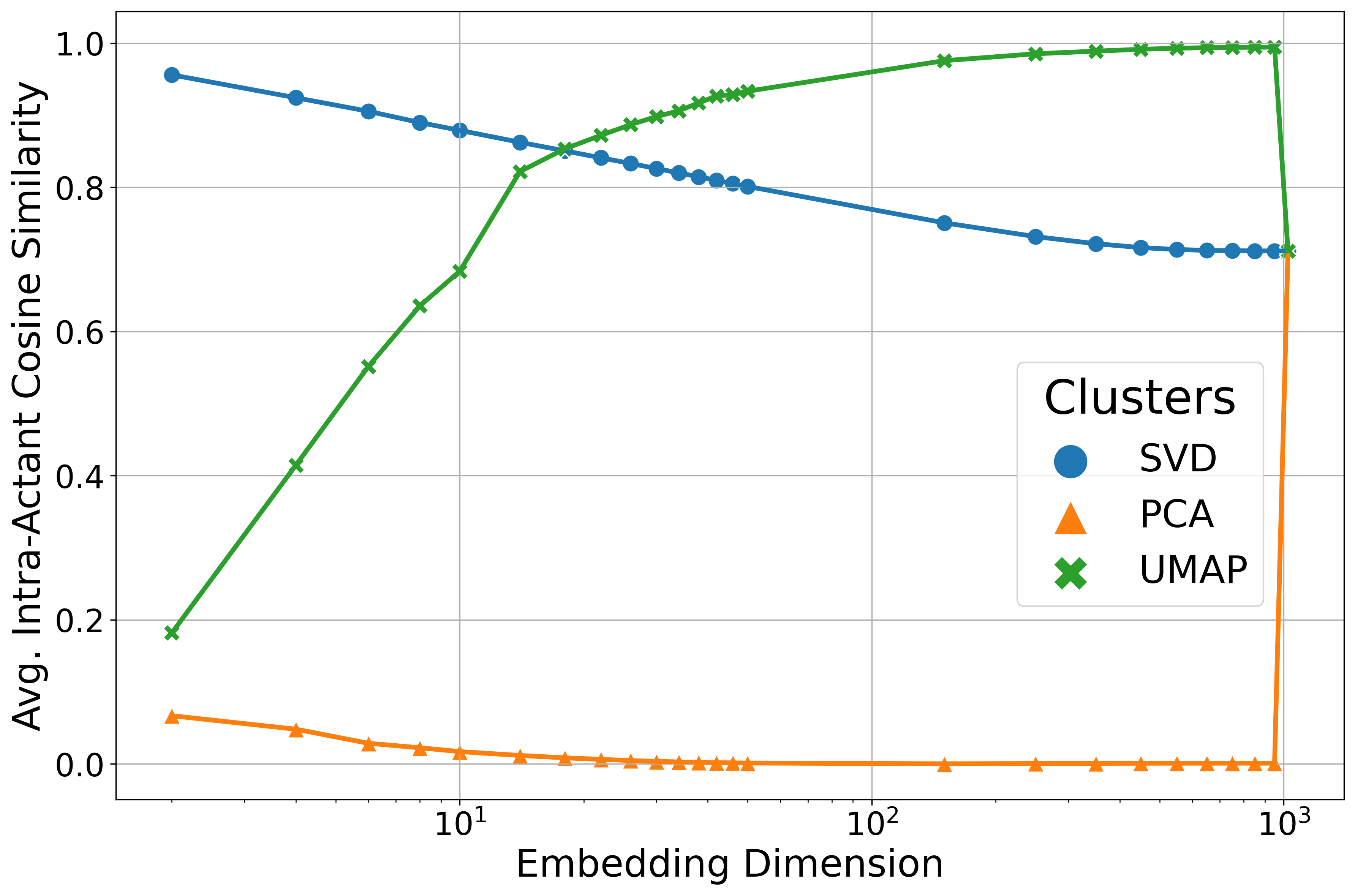}
    \caption{Average actant cosine similarity for whole actant corpus over different levels of dimension reduction. The starting point is \textit{E5-Large} embedding with dimension $1024$.}
    \label{fig:actant_similarity_on_dimension}
\end{figure}

First, we monitor the change in similarity between all actants in the corpus. This involves calculating the similarity of each actant with every other actant. Specifically, we compute the average of the sub-diagonal elements of the similarity matrix for all actants. Figure \ref{fig:actant_similarity_on_dimension} shows this value for different dimensions and techniques, i.e., UMAP, SVD, and PCA. Each iteration is calculated from the full embedding (rightmost observation).

Each of the methods shows a unique behaviour. PCA completely nullifies every structure in the embeddings. The similarity values close to zero suggest a random distribution of vectors. SVD shows an expected behaviour, with the similarity gradually increasing with decreasing dimension. The only difference between SVD and PCA is the centring of the variable for the latter. Thus the extreme behaviour of PCA is likely caused by this initial step.

UMAP goes a different route. Initially, the similarity drastically inflates to a value close to $1$ and then slowly degrades. We hypothesise, that the projection to a lower dimensional space causes this behaviour. In contrast to the previous methods, UMAP is not linear. Instead of combining the main features to find the axis of highest variance, UMAP actively \textit{moves} the different points. If all points are moved towards a lower-dimensional subspace of the original space, this would align the embeddings and thus explain the inflation of the similarity score. Only in lower dimensions, do we recover some structure.  

\subsubsection*{Individual similarity}
\begin{figure}
    \centering
    \includegraphics[width=0.5\textwidth]{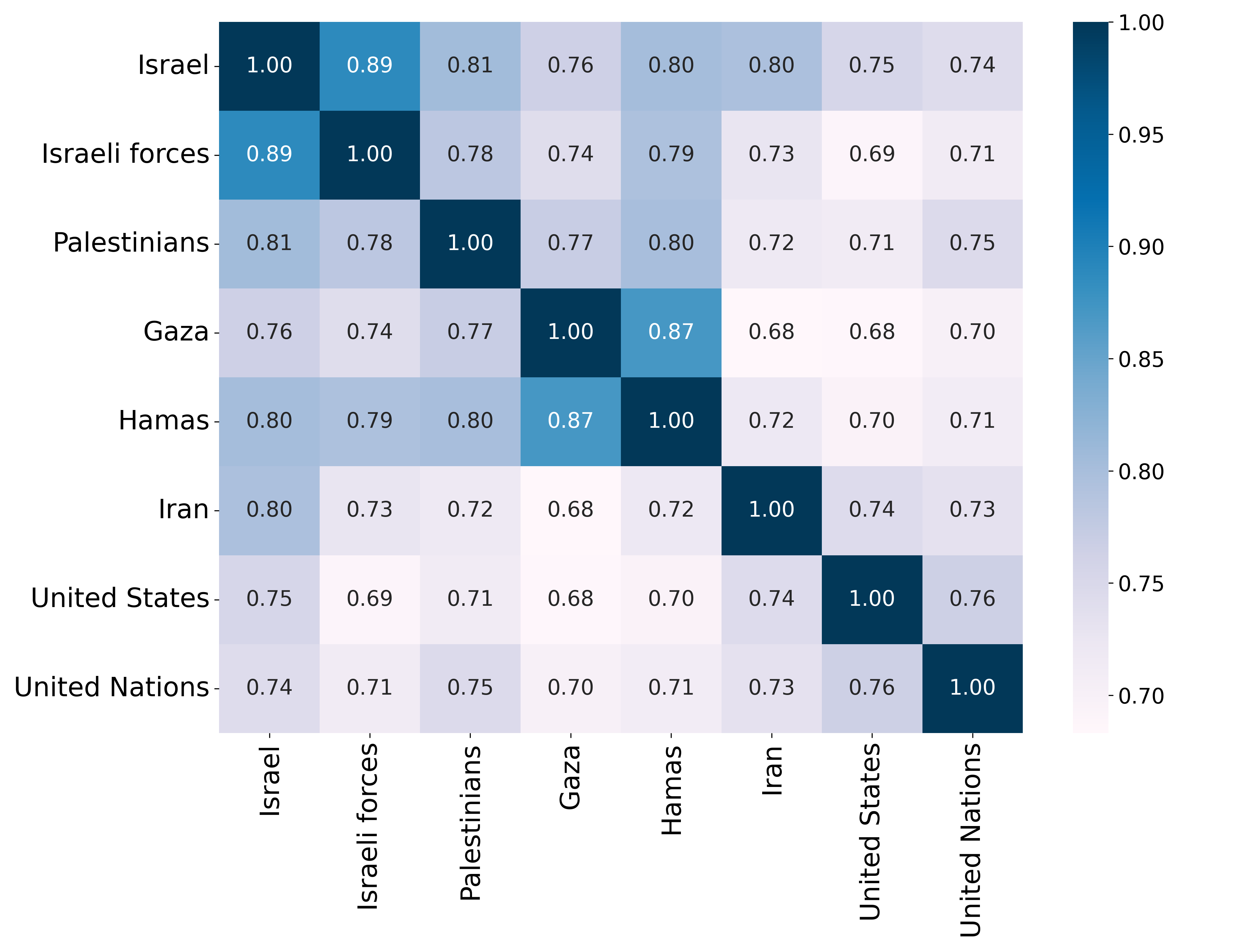}
    \caption{Cosine similarity of key actors within the dataset using \textit{E5-Large} embeddings.}
    \label{fig:actant_similarity_matrix}
\end{figure}

\begin{figure}
    \centering
    \includegraphics[width=0.5\textwidth]{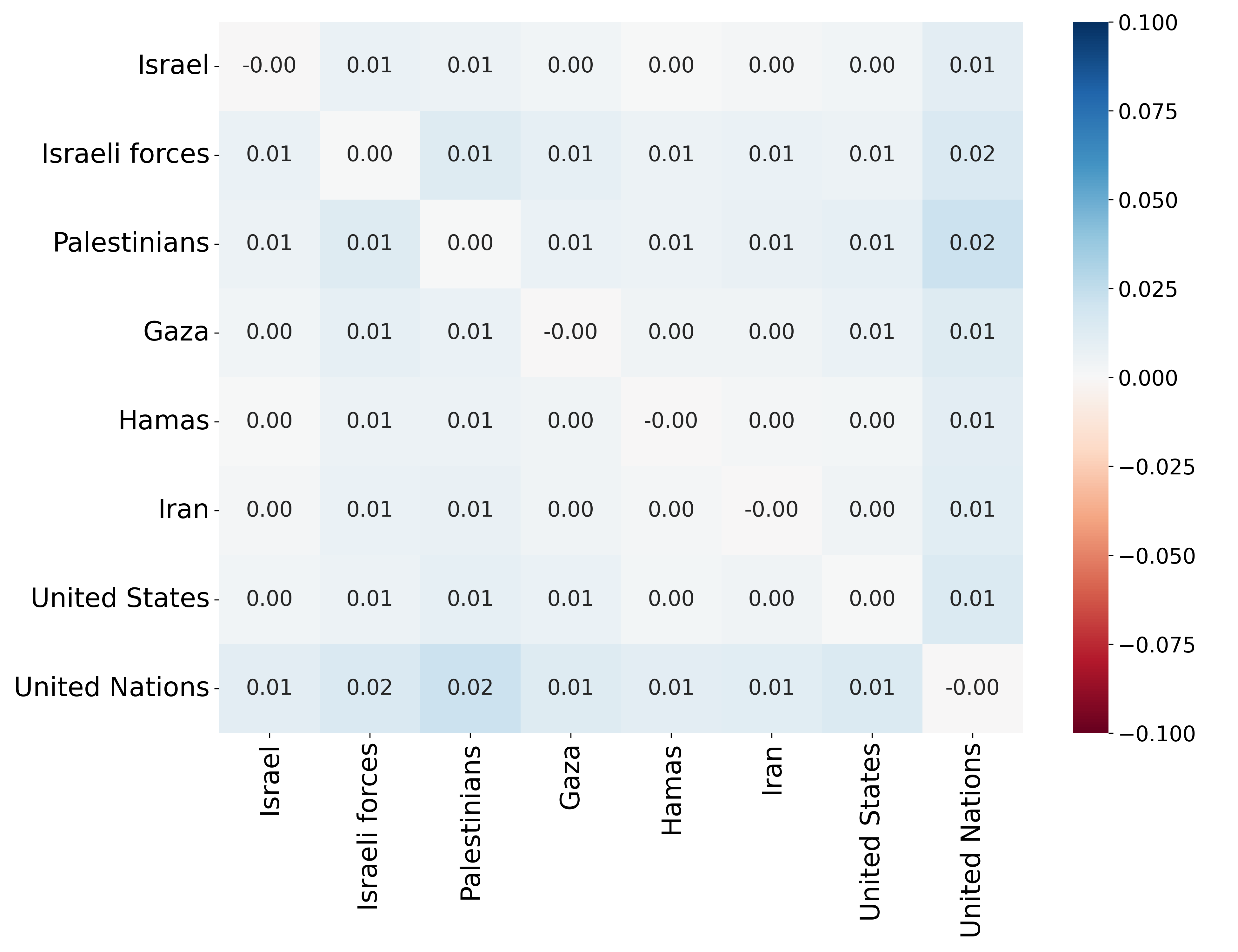}
    \caption{Difference in cosine similarity between the \textit{SVD-34} and the full \textit{E5-Large} embedding.}
    \label{fig:actant_similarity_matrix_diff}
\end{figure}

From the previous experiment, we can discard both UMAP and PCA. In the following, we assess, how SVD impacts the similarity of core actors in our dataset. This serves as a qualitative insight into the latent semantic similarity space. We compare the cosine similarity for the original embedding with a dimension of 1024 \ref{fig:actant_similarity_matrix}. Next, we observe the change in similarity between the full dimension and the reduced embeddings with a dimension of 34 \ref{fig:actant_similarity_matrix_diff}. We chose this dimension to achieve high dimension reduction, without too much inflation in the average similarity between actors.

Figure \ref{fig:actant_similarity_matrix} showcases the change of actant similarities through these methods. On the left, we can see the \textit{orignal} similarities, i.e., the cosine similarities of the full \textit{E5-Large} embeddings, of some of the main actors. Figure \ref{fig:actant_similarity_matrix_diff} shows the difference between these original embeddings and the ones that have been reduced using SVD. The differences in similarities before and after dimension reduction are small.

% \subsection{Metrics}
% \begin{itemize}
%     \item AMI
%     \item Intra-Cluster Similarity (ICS)
%     \item Actant Similarity
% \end{itemize}

\subsection{Additional Figures and Tables}

\begin{figure}
    \centering
    \includegraphics[width=0.475\textwidth]{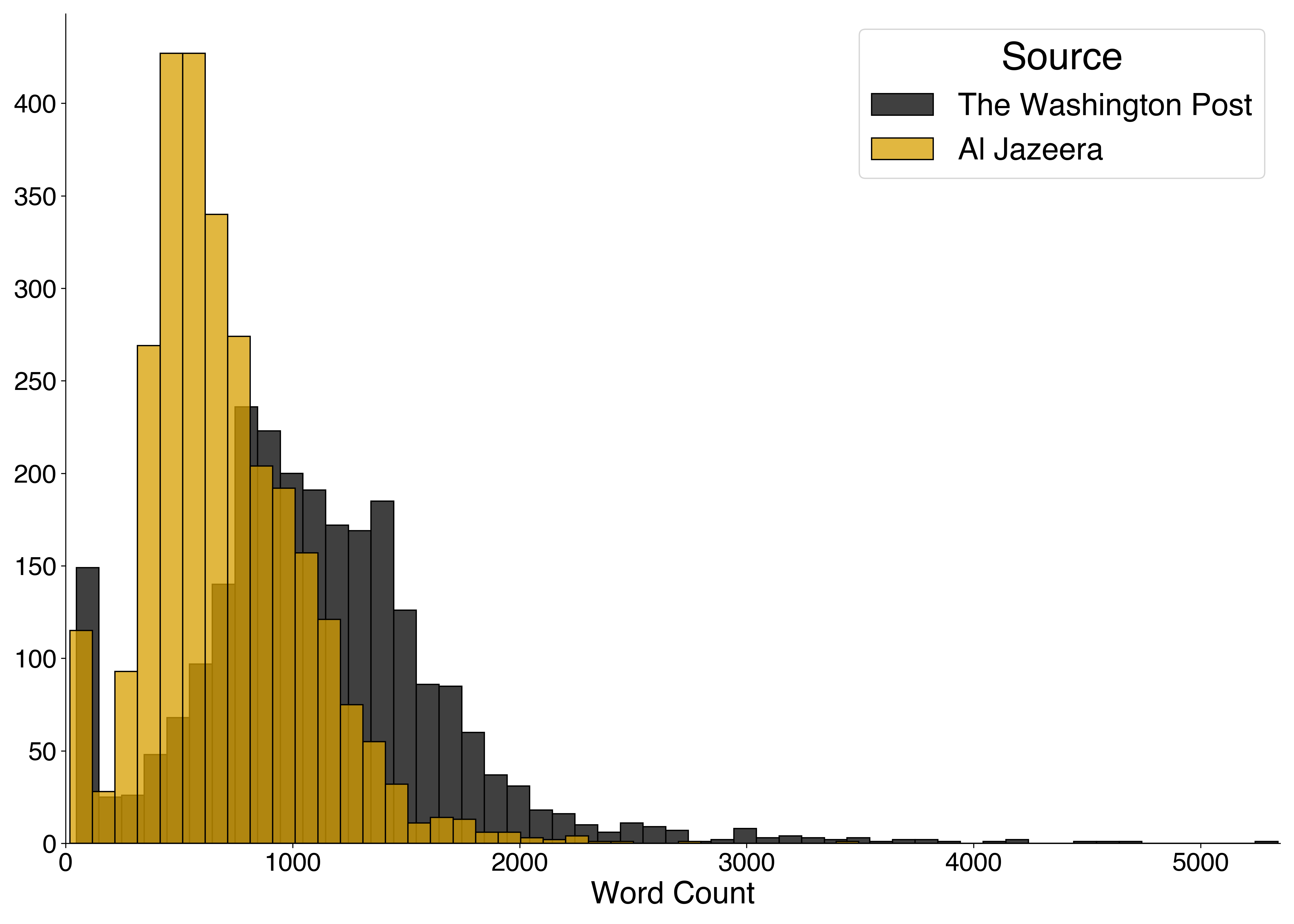}
    \caption{Distribution of word count per article for Al Jazeera and The Washington Post.}
    \label{fig:word_count}
\end{figure}

\begin{figure*}
    \centering
    \makebox[\textwidth][c]{\includegraphics[width=1\textwidth]{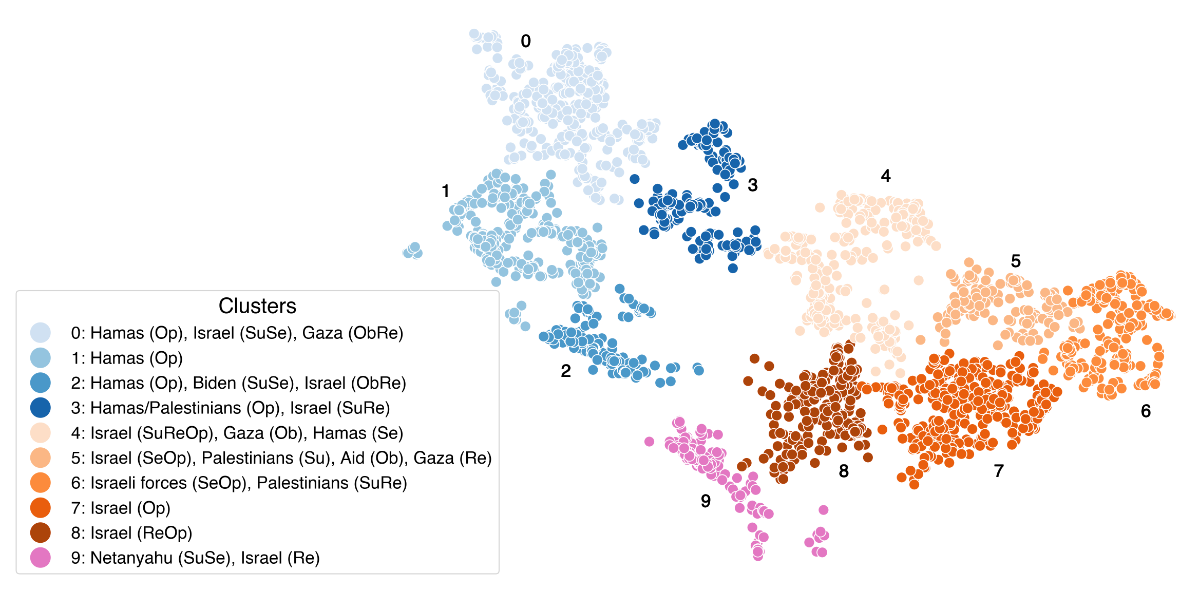}}%
    \caption{Map of 10 identified narratives directly related to the Israel-Palestine conflict.}
    \label{fig:narrative_map_top_half}
\end{figure*}

\begin{figure*}
    \centering
    \makebox[\textwidth][c]{\includegraphics[width=1\textwidth]{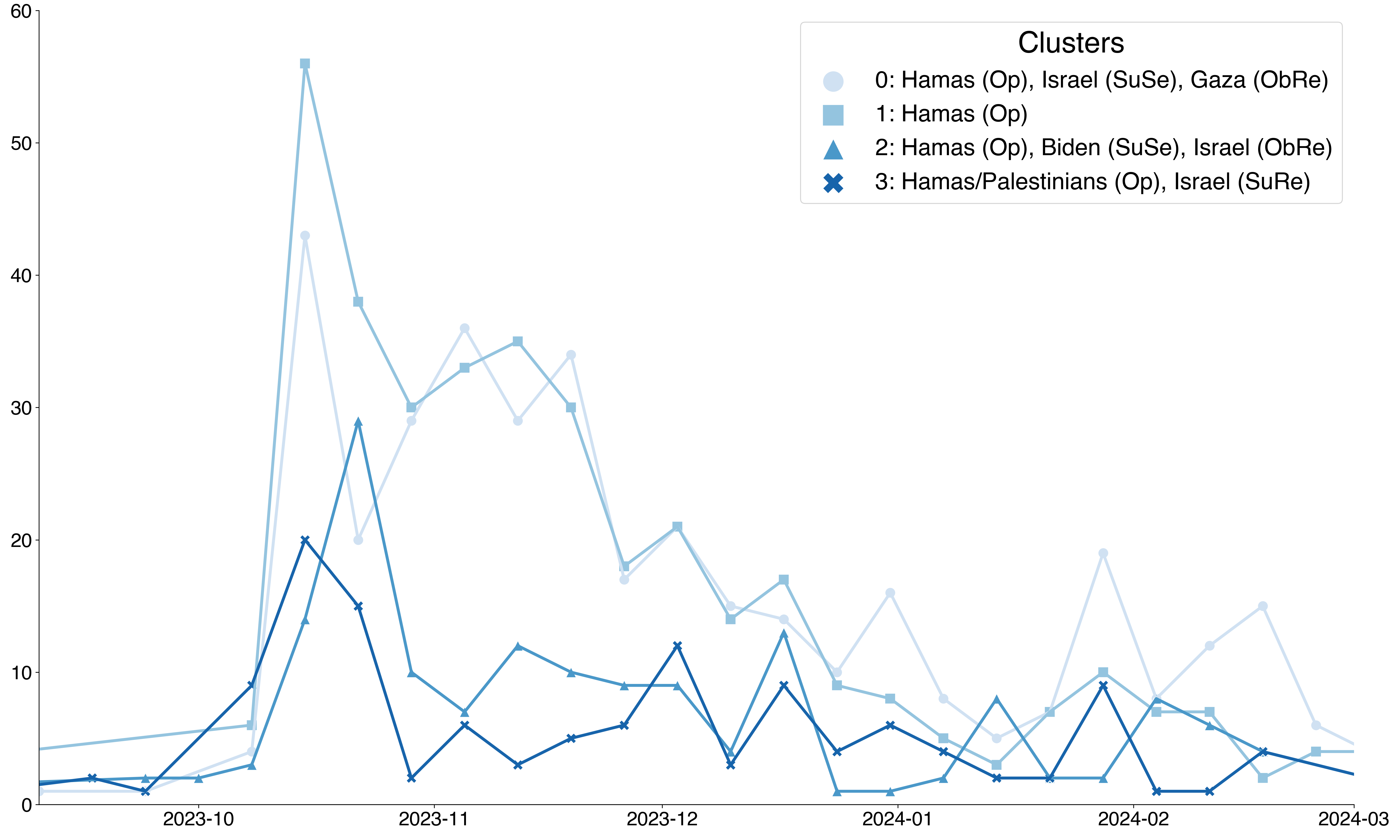}}%
    \caption{Number of articles per week between September 2023, and March 2024 that are part of the component ``Hamas as Opponent''.}
    \label{fig:timeline_clusters_hamas_opponent}
\end{figure*}

\begin{figure*}
    \centering
    \makebox[\textwidth][c]{\includegraphics[width=1\textwidth]{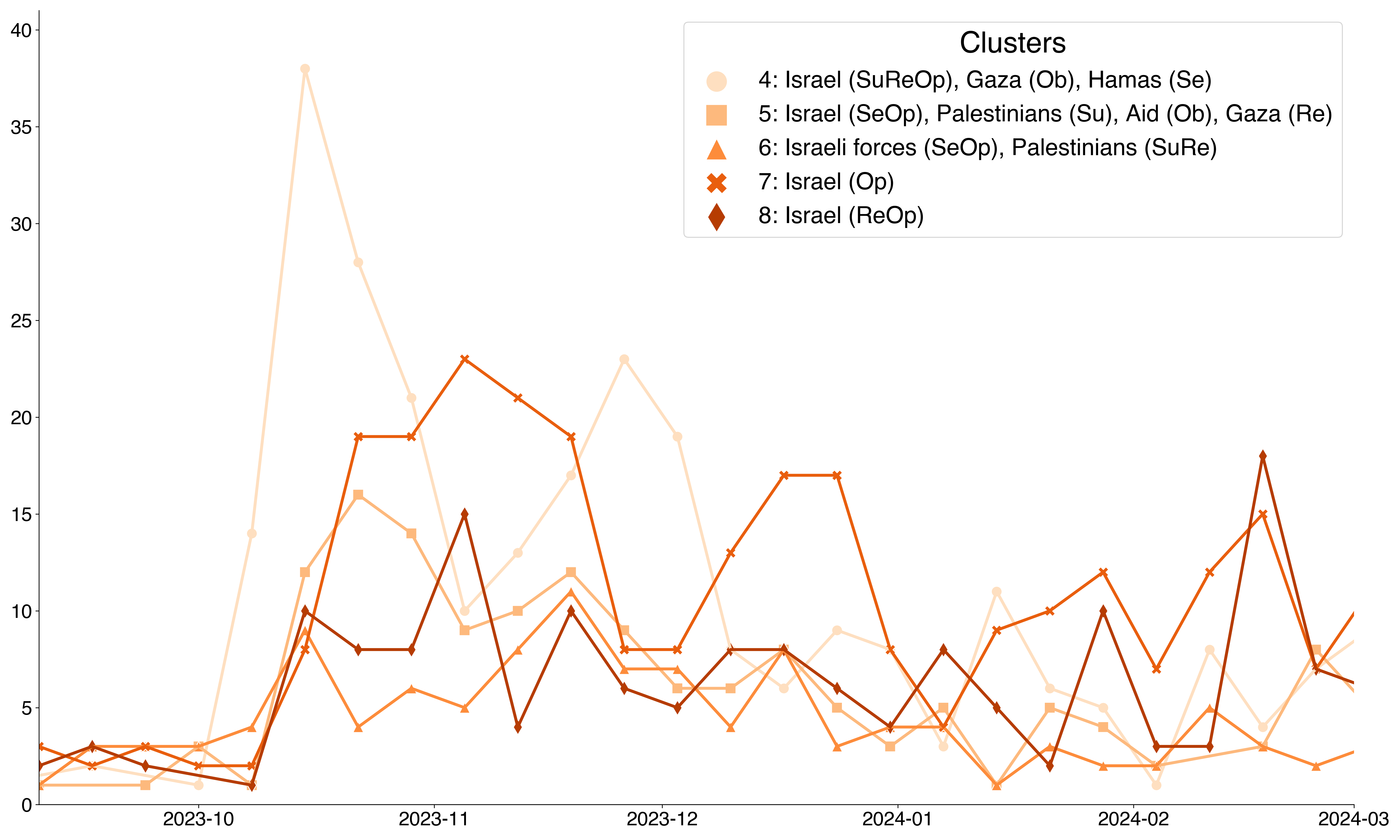}}%
    \caption{Number of articles per week between September 2023, and March 2024 that are part of the component ``Israel as Opponent''.}
    \label{fig:timeline_clusters_israel_opponent}
\end{figure*}

\begin{table}
\caption{Top 3 actors per actant for both Al Jazeera and The Washington Post. Numbers in braces show the number of articles for each actor.}

\centering
\begin{tabular}{lll}
\toprule
 & Al Jazeera (2872) & The Washington Post (2470) \\
\midrule
Subject & \makecell[lt]{ Israel (475)\\Benjamin Netanyahu (73)\\Joe Biden (63) \\} & \makecell[lt]{ Israel (239)\\President Biden (173)\\The author (62) \\} \\
& & \\
Object & \makecell[lt]{ Gaza (134)\\Israel (91)\\Gaza Strip (82) \\} & \makecell[lt]{ Gaza (113)\\Israel (111)\\Ukraine (47) \\} \\
& & \\
Sender & \makecell[lt]{ Israel (291)\\Israeli forces (105)\\Hamas (74) \\} & \makecell[lt]{ President Biden (137)\\Israel (108)\\Hamas (84) \\} \\
& & \\
Receiver & \makecell[lt]{ Israel (369)\\Palestinians (149)\\Gaza (135) \\} & \makecell[lt]{ Israel (250)\\Hamas (66)\\Gaza (65) \\} \\
& & \\
Helper & \makecell[lt]{ Qatar (56)\\Egypt (52)\\United States (36) \\} & \makecell[lt]{ United States (51)\\U.S. officials (36)\\Qatar (35) \\} \\
& & \\
Opponent & \makecell[lt]{ Israel (768)\\Hamas (368)\\Israeli forces (139) \\} & \makecell[lt]{ Hamas (536)\\Israel (180)\\Russia (80) \\} \\
& & \\
\bottomrule
\end{tabular}

\label{tab:top_3_actors}
\end{table}

\onecolumn
% \small
\begin{landscape}
\footnotesize

% \begin{sidewaystable}
\begin{longtable}{|l|l|l|l|l|l|l|}
\caption{Most common actors per actant for each cluster. We only include actors that occur in at least \SI{5}{\percent} of articles and a maximum of three per cluster.}

\label{tab:clusters}
\\
    
% \begin{tabular}{lllllll}
\toprule
 & Subject & Object & Sender & Receiver & Helper & Opponent \\
Cluster &  &  &  &  &  &  \\
\midrule
\endfirsthead

\toprule
& Subject & Object & Sender & Receiver & Helper & Opponent \\
 &  &  &  &  &  &  \\
\midrule
\endhead

\bottomrule
\endfoot

\bottomrule
\endlastfoot
% \midrule
\rowcolor{gray!25}
0 & $66\%$: Israel & $28\%$: Gaza & $37\%$: Israel & $23\%$: Gaza &  & $88\%$: Hamas \\
\rowcolor{gray!25}
 & $9\%$: Benjamin Netanyahu & $12\%$: Gaza Strip & $7\%$: Israel Defense Forces & $20\%$: Hamas & $7\%$: United States &  \\
\rowcolor{gray!25}
 &  & $5\%$: Hamas &  & $10\%$: Palestinians & $6\%$: Egypt &  \\[0.3cm]

1 &  & $13\%$: Israel & $8\%$: Hamas & $16\%$: Israel &  & $91\%$: Hamas \\
 &  &  &  &  &  & $5\%$: Hamas militants \\[0.3cm]
 
\rowcolor{gray!25}
2 & $33\%$: President Biden & $25\%$: Israel & $31\%$: President Biden & $49\%$: Israel &  & $65\%$: Hamas \\
\rowcolor{gray!25}
 & $18\%$: Joe Biden &  & $15\%$: Joe Biden &  & $8\%$: U.S. officials & $9\%$: Israel \\
\rowcolor{gray!25}
 & $15\%$: Antony Blinken &  & $12\%$: Antony Blinken &  &  &  \\[0.3cm]

3 & $62\%$: Israel & $11\%$: Gaza & $16\%$: Hamas & $49\%$: Israel &  & $37\%$: Hamas \\
 &  & $9\%$: Palestinians & $8\%$: Israel & $11\%$: Palestinians & $6\%$: United States & $7\%$: Palestinians \\
 &  & $7\%$: Gaza Strip & $6\%$: United States &  &  &  \\[0.3cm]
 
\rowcolor{gray!25} 
4 & $57\%$: Israel & $17\%$: Gaza & $23\%$: Hamas & $38\%$: Israel &  & $83\%$: Israel \\
\rowcolor{gray!25}
 & $20\%$: Hamas & $9\%$: Gaza Strip & $15\%$: Israel & $12\%$: Gaza & $12\%$: Qatar &  \\
\rowcolor{gray!25}
 & $5\%$: Palestinians & $7\%$: Israel & $6\%$: Qatar & $9\%$: Palestinians & $7\%$: Egypt &  \\[0.3cm]

5 & $8\%$: Palestinians & $8\%$: aid & $27\%$: Israel & $17\%$: Gaza Strip &  & $80\%$: Israel \\
 &  & $5\%$: Gaza &  & $16\%$: Palestinians & $8\%$: United Nations &  \\
 &  &  &  & $15\%$: Gaza & $5\%$: UNRWA &  \\[0.3cm]
 
\rowcolor{gray!25}
6 & $8\%$: Palestinians &  & $27\%$: Israeli forces & $13\%$: Palestinians &  & $28\%$: Israeli forces \\
\rowcolor{gray!25} 
 &  &  & $12\%$: Israel &  &  & $15\%$: Israel \\
\rowcolor{gray!25} 
 &  &  & $12\%$: Israeli army &  &  & $12\%$: Israeli army \\[0.3cm]

7 & $5\%$: Shireen Abu Akleh &  &  &  &  & $68\%$: Israel \\
 &  &  &  &  &  & $10\%$: Israeli forces \\[0.3cm]
 
\rowcolor{gray!25} 
8 & $6\%$: Antony Blinken & $9\%$: Israel &  & $60\%$: Israel &  & $54\%$: Israel \\[0.3cm]

% 9 & $44\%$: Benjamin Netanyahu & $6\%$: judicial overhaul & $27\%$: Benjamin Netanyahu & $12\%$: Israelis &  & $7\%$: \makecell[lt]{Netanyahu's \\government} \\
9 & $44\%$: Benjamin Netanyahu & $6\%$: judicial overhaul & $27\%$: Benjamin Netanyahu & $12\%$: Israelis &  & $7\%$: Netanyahu's government \\
 % & $15\%$: Protesters & $5\%$: judicial system & $7\%$: \makecell[lt]{ Prime Minister \\Benjamin Netanyahu} & $12\%$: Israel &  & $5\%$: Palestinians \\
 & $15\%$: Protesters & $5\%$: judicial system & $7\%$: Prime Minister Benjamin Netanyahu & $12\%$: Israel &  & $5\%$: Palestinians \\
 & $14\%$: Netanyahu &  & $6\%$: Netanyahu &  &  &  \\[0.3cm]

\rowcolor{gray!25} 
10 &  &  &  &  &  &  \\[0.3cm]
 
11 & $12\%$: Iran &  & $10\%$: United States & $14\%$: Iran &  & $23\%$: Iran \\
 & $10\%$: United States &  & $8\%$: Iran & $8\%$: Israel &  & $13\%$: Israel \\
 & $10\%$: Houthis &  & $5\%$: Houthis & $7\%$: Houthis &  & $11\%$: Houthis \\[0.3cm]

\rowcolor{gray!25} 
12 & $13\%$: Ukraine & $26\%$: Ukraine & $6\%$: President Biden & $33\%$: Ukraine &  & $68\%$: Russia \\
\rowcolor{gray!25} 
 &  &  &  & $5\%$: Russia &  &  \\[0.3cm]
 
13 & $43\%$: President Biden &  & $29\%$: President Biden & $12\%$: Congress &  & $11\%$: China \\
 & $10\%$: Joe Biden &  & $9\%$: Biden administration & $10\%$: President Biden &  & $10\%$: Donald Trump \\
 & $10\%$: Biden &  &  &  &  & $9\%$: Trump \\[0.3cm]

\rowcolor{gray!25} 
14 & $10\%$: House Republicans & $9\%$: speaker & $11\%$: House Republicans & $17\%$: House Republicans &  & $7\%$: President Biden \\
\rowcolor{gray!25}
 & $8\%$: Mike Johnson &  &  & $8\%$: Congress &  & $5\%$: House Republicans \\
\rowcolor{gray!25} 
 & $7\%$: Kevin McCarthy &  &  & $8\%$: President Biden &  &  \\[0.3cm]
 
15 & $36\%$: Donald Trump &  & $19\%$: Donald Trump &  &  & $32\%$: Donald Trump \\
 & $10\%$: Nikki Haley &  & $6\%$: Nikki Haley &  &  & $26\%$: Trump \\
 & $6\%$: Ron DeSantis &  &  &  &  &  \\[0.3cm]

\rowcolor{gray!25} 
16 & $19\%$: The author &  & $9\%$: the author & $9\%$: the reader &  &  \\
\rowcolor{gray!25} 
 &  &  & $9\%$: The author &  &  &  \\[0.3cm]
 
17 & $36\%$: Israel &  & $26\%$: Israel & $11\%$: Syria &  &  \\
 &  &  &  & $7\%$: Gaza &  & $11\%$: Iran \\
 &  &  &  &  &  & $7\%$: Hezbollah \\[0.3cm]

% \bottomrule
% \end{tabular}

\end{longtable}
\end{landscape}
\twocolumn
\normalsize
% \end{sidewaystable}
\begin{table}[t]
\caption{Shares and total number of articles for the eight most common actant syncretisms in the data.}
    \centering
\begin{tabular}{lrr}
\toprule
Syncretism & Share & Total \\
\midrule
Subject-Sender     &	0.41 &	2195 \\
Subject-Receiver   &	0.16 &	866 \\
Object-Receiver    &	0.16 &	853 \\
Opponent-Receiver  &	0.14 &	758 \\
Opponent-Sender    &	0.09 &	499 \\
Opponent-Subject   &	0.05 &	260 \\
Opponent-Object    &	0.05 &	241 \\
Sender-Helper      &	0.02 &	101 \\
\bottomrule
\end{tabular}

    \label{tab:syncretism}
\end{table}
\begin{table}
\caption{Shares of Al Jazeera and The Washington Post articles for each cluster.}

    \centering
    
\begin{tabular}{lrr}
\toprule
Cluster & Al Jazeera & The Washington Post \\
\midrule
0 & 0.53 & 0.47 \\
1 & 0.29 & 0.71 \\
2 & 0.33 & 0.67 \\
3 & 0.65 & 0.35 \\
4 & 0.76 & 0.24 \\
5 & 0.83 & 0.17 \\
6 & 0.91 & 0.09 \\
7 & 0.81 & 0.19 \\
8 & 0.78 & 0.22 \\
9 & 0.70 & 0.30 \\
10 & 0.39 & 0.61 \\
11 & 0.70 & 0.30 \\
12 & 0.44 & 0.56 \\
13 & 0.19 & 0.81 \\
14 & 0.17 & 0.83 \\
15 & 0.13 & 0.87 \\
16 & 0.19 & 0.81 \\
17 & 0.61 & 0.39 \\
\bottomrule
\end{tabular}
    \label{tab:source_shares}
\end{table}

% - Number of actors extracted per actant
% - Media cloud links

% \section{Algortihmic specifications}
% \begin{itemize}
%     \item SVD algorithm
%     \item Clustering algorithm

% \end{itemize}

% \subsection{Technology}
% \begin{itemize}
%     \item Jupyter
%     \item Huggingface

% \end{itemize}

\end{document}